\documentclass{article}

\usepackage{arxiv}

\usepackage{subfiles}
\usepackage{caption}
\usepackage{subcaption}
\usepackage{graphicx}
\usepackage{comment}

\usepackage[utf8]{inputenc} 
\usepackage[T1]{fontenc}    
\usepackage{hyperref}       
\usepackage{url}            
\usepackage{booktabs}       
\usepackage{amsfonts}       

\title{Predicting and Understanding Human Action Decisions during Skillful Joint-Action via Machine Learning and Explainable-AI}

\author{
 Fabrizia Auletta \\
  School of Psychological Sciences, Faculty of Medicine, Health and Human Sciences\\
  Macquarie University\\
  Sydney, NSW, Australia\\
  \texttt{fabrizia.auletta@hdr.mq.edu.au} \\
   \And
 Rachel W. Kallen \\
  Center for Elite Performance, Expertise and Training\\
  Macquarie University\\
  Sydney, NSW, Australia\\
  \texttt{rachel.kallen@mq.edu.au} \\
  \And
 Mario di Bernardo \\
  Department of Electrical Engineering and Information Technology\\
  University of Naples Federico II\\
  Naples, Italy\\
  \texttt{mario.dibernardo@unina.it} \\
  \And
 Micheal J. Richardson \\
  Center for Elite Performance, Expertise and Training\\
  Macquarie University\\
  Sydney, NSW, Australia\\
  \texttt{michael.j.richardson@mq.edu.au} \\
}

\begin{document}
\maketitle
\begin{abstract}
This study uses supervised machine learning (SML) and explainable artificial intelligence (AI) to model, predict and understand human decision-making during skillful joint-action. Long short-term memory networks were trained to predict the target selection decisions of expert and novice actors completing a dyadic herding task. Results revealed that the trained models were expertise specific and could not only accurately predict the target selection decisions of expert and novice herders but could do so at timescales that preceded an actor’s conscious intent. To understand what differentiated the target selection decisions of expert and novice actors, we then employed the explainable-AI technique, SHapley Additive exPlanation, to identify the importance of informational features (variables) on model predictions. This analysis revealed that experts were more influenced by information about the state of their co-herders compared to novices. The utility of employing SML and explainable-AI techniques for investigating human decision-making is discussed.
\end{abstract}

\keywords{decision-making, supervised machine learning, explainable-AI, artificial neural networks, joint-action, multi-agent interaction}

\section{Introduction}\label{introduction}

Performing tasks with other individuals is an essential part of everyday human life. Such behavior requires that co-actors reciprocally coordinate their actions with respect to each other and changing task demands \cite{dale2013self, richardson2016symmetry, sebanz2021progress, schmidt2011understanding}. Pivotal to the structural organization of such action is the ability of co-actors to effectively decide how and when to act \cite{sebanz2009prediction}, with robust decision-making often differentiating expert from non-expert performance \cite{davids2015expert}. This is true whether one considers the simple activity of family members loading a dishwasher together \cite{richardson2007judging, sebanz2006joint, vesper2017joint}, or the more complex activities performed by elite athletes during team sports \cite{araujo2006ecological, yamamoto2013joint} or soldiers during high-stakes military operations \cite{worm1998joint}. 

In contrast to practical reasoning or deliberative decision-making, where an actor extensively evaluates all possibilities to determine the optimal action, the decision-making that occurs during skillful action is typically fast-paced and highly context dependent \cite{turvey2007action, wolpert2012motor, goldstone2009collective, spivey2006continuous}, with actors spontaneously adapting their actions to achieve task goals as ``best as possible'' \cite{christensen2016cognition}. Indeed, the effectiveness of action decisions during skillful behavior is a function of an actor’s level of situational awareness \cite{christensen2019memory, martens2020doing}, with task expertise reflecting the attunement of an actor to the information that specifies what action possibilities (optimal or sub-optimal) ensure task completion \cite{jacobs2007direct, van2015information, zhao2015line}. Developing a comprehensive understanding of decision-making during skillful behavior therefore rests on the ability of researchers and practitioners to identify what task information underlies effective decision-making performance. Key to achieving this, is developing decision-making models that can not only predict the action decisions of human actors during skillful action, but also help to identify what differentiates expert from non-expert performance. Motivated by these challenges, the current study proposes the use of state-of-the art Supervised Machine Learning (SML) and explainable AI (Artificial Intelligence) techniques to model, predict and explicate the action decisions of expert and novice pairs performing a fast-paced joint-action task. Specifically, for the first time in the literature, we show how using these computational techniques renders it possible to uncover the crucial information that co-actors exploit when making action decisions in  joint-action (and individual) task contexts.

\subsection{Supervised Machine Learning}
The application of Machine Learning (ML) techniques has rapidly increased over the last decade. For example, ML is now integral to modern image and speech recognition \cite{simonyan2014very, he2016deep, deng2013machine, amodei2016deep}, scientific analysis \cite{hamadeh2020machine,krause2018grader}, financial modeling \cite{lin2011machine}, and online product, movie and social interest recommendations \cite{boukerche2020machine, tuinhof2018image, park2009individual}. In many contexts, ML models are trained via SML, whereby computational models learn to correctly classify input data, or predict future outcomes states from input data, by leveraging coded, real-word training samples \cite{goodfellow2016machine, langley1996elements}. Training samples include representative task data (e.g. images, sounds, motion data) that have been labeled with the correct data class or outcome state. These training samples are then used to build an approximate model of how the input data (e.g., pixels from an image) map to the correct output label (e.g., cat or dog) \cite{hastie2009overview, caruana2006empirical}. Following training, the efficacy of the model is then tested against data not supplied during training, with effective models able to generalize the learned input-output mappings to unseen data.  
 
\subsection{Artificial Neural and Long Short-Term Memory Networks}
SML models can be realized in numerous ways, for instance, using decision trees \cite{quinlan1986induction, safavian1991survey}, support vector machines \cite{cortes1995support, cristianini2000introduction} or, of particular importance here, Artificial Neural Networks (ANNs). In general, ANNs are a composition of elementary units, \textit{nodes}, that are grouped in interconnected \textit{layers}, where nodes have different activation biases and the connections between nodes can have different weights. A typical ANN includes an input and an output layer, with 1 or more ``hidden layers'' in between (with deeper ANNs having more hidden layers). Training an ANN to model input-output mappings via SML requires finding the combination of network parameters (weights and biases) that map input data to the correct output class or state prediction. This is achieved by iteratively evaluating the error between the correct and predicted output of the ANN (via a \textit{loss function}) and adjusting the network parameters to minimize this error using a process called \textit{back-propagation}  (see e.g., \cite{basheer2000artificial} for more details).

There are various types of ANNs. Of relevance here, is an ANN known as a \textit{Long Short-Term Memory} (LSTM) network, which is a form of recurrent neural network that in addition to feed-forward connections among node layers also includes feedback connections. These feedback connections enable the ANN to process and retain information about sequences of consecutive inputs \cite{lstm_1997,hinton2012improving}. Accordingly, LSTMs are commonly used in time-series prediction tasks \cite{gers2002applying,chimmula2020time,alhagry2017emotion, wang2016attention,naretto2020prediction}, where the processing of both past and present input states is required to correctly predict future states. LSTMs are applicable here, as human decision-making is based on the assessment of dynamical (time varying) task information \cite{zhao2015line, araujo2006ecological} and, thus, the prediction of future state decisions requires processing sequences of task relevant state input data.   
	
\subsection{Explainable AI}
Despite the increasing utility and effectiveness of ANNs in recent years \cite{he2016deep,deng2013machine,tuinhof2018image,angelini2008neural}, the large number of connection weights within ANNs, particularly Deep-ANNs, makes it difficult to directly access how input features relate to output predictions. For this reason, ANNs are often referred to as ``black-box'' models. However, a desire to better understand and interpret the validity of ANNs and other black-box models, as well as the growing demand for more ethical and transparent AI systems \cite{Voigt2017gdpr}, has resulted in a renewed interest in the application and development of explainable-AI techniques \cite{lundberg2018explainable,parsa2020toward,slack2020fooling,mokhtari2019interpreting,naretto2020prediction} such as LIME \cite{ribeiro2016should}, DeepLIFT \cite{shrikumar2016not}, and, more recently, SHapley Additive exPlanation (SHAP) \cite{lundberg2017unified, lundberg2020local}, which we employ here. 

These techniques make the internal processes of a black-box model understandable by deriving linear explanation functions of the effects that input features have on output states. For example, the SHAP algorithm pairs each input feature with a SHAP value. The higher the SHAP value, the greater the influence that feature has on an output state. Given that SHAP values are locally accurate, one can derive a measure of global feature importance by calculating the average importance of a feature over the test set used to assess model accuracy. The result is an average SHAP value for a given input to output mapping that captures the overall significance of a given input feature for a given output prediction.
	
\subsection{Current Study}\label{sec:CurrentStudy}
The current study had three primary aims: (1) investigate the use of SML trained LSTM-layered ANN models (hereafter referred to LSTM$_{NN}$ models) to predict human decision-making during skillful joint-action;  (2) demonstrate how SHAP can be employed to uncover the task information that supports human decision-making during skillful joint-action by determining what input information (features) most influenced the output predictions of a trained LSTM$_{NN}$ model; and (3) apply these techniques to explicate differences between the decision-making process of novice and expert players while playing a simulated, fast paced herding game \cite{richardson_modeling_2016, nalepka_herd_2017, Nalepka_2019}. 


Herding tasks involve the interaction of two sets of autonomous agents — one or more \textit{herder} agents are required to corral and contain a set of heterogeneous \textit{target} agents. Such activities are ubiquitous in daily life and provide a prototypical example of everyday skillful joint- or multiagent behavior. Indeed, while the most obvious examples involve farmers herding sheep or cattle, similar task dynamics define teachers corralling a group of young children through a museum or firefighters evacuating a crowd of people from a building \cite{ma2016effective}.

 \begin{figure}[tbhp]
 	\centering
 	\includegraphics[width=\linewidth]{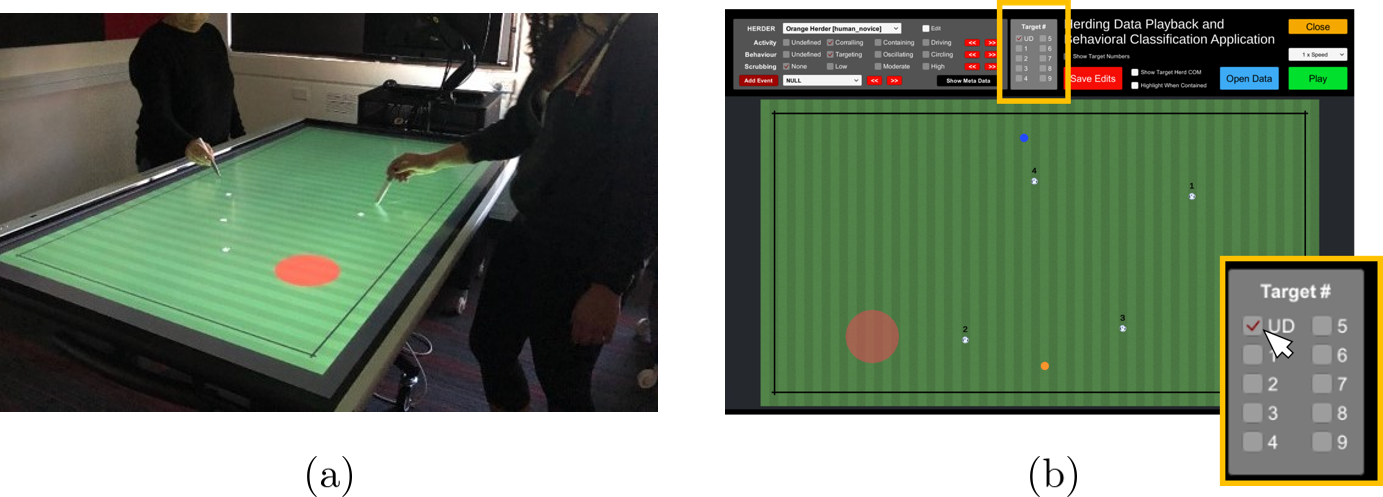}
 	\caption{(a) The herding task and experimental setup. (b) A screenshot of the data playback and coding application used to identify which target a herder was corralling at each time step. The blue and orange dots are the herders and the white dots are the targets. The text label above each target specifies the target number (i.e, 1 to 4). The transparent red area is the containment region, which for the data employed here was always in the center of the game field. The insert panel on the bottom right of (b) is the target coding panel, where UD = no target. See main text for more details.}
 	\label{fig:experiment}
 \end{figure} 
 
For the current study, we modeled data from \cite{Rigoli_AAMAS_2020}, in which pairs of players controlled virtual herder agents to corral a herd of four virtual cows (hereafter refereed to as \textit{targets}), dispersed around a game field, into a red containment area positioned at the center of the game field. The task was presented on a large touch screen, with players using a touch-pen stylus to control their virtual herders (Figure~\ref{fig:experiment}(a)). Targets were repelled away from the human-controlled herders when the herder came within a certain distance of the target. When not influenced by a herder, the movement of targets was governed by Brownian motion, with targets randomly (diffusely) wandering around the game field.

To successfully corral the targets into the containment area, previous research has demonstrated that effective task performance requires that players coordinate their target selection decisions by dynamically dividing the task space into two regions of responsibility, with each player then corralling one target at a time in their corresponding regions of responsibility until all of the targets are corralled within the containment area \cite{nalepka_herd_2017, Rigoli_AAMAS_2020}. To date, however, no previous research has explicitly explored or modeled the process by which players make their target selection decisions, nor what state information regulates this decision-making behavior. Assuming that a player's (a) target selection decisions are specified in the movements \cite{becchio2014kinematic, cavallo2016decoding, runeson1983kinematic}  of their herder and (b) that players used kinematic information of target and herder states (e.g., relative positions, velocities) to make target selection decisions \cite{nalepka_herd_2017, jacobs2007direct, van2015information, zhao2015line}, we expected that an LSTM$_{NN}$ could be trained via SML to predict these target selection decisions using short sequences of herder and target state input features that preceded target selection. We expected that target selection decisions could be predicted prior to a player enacting a target selection decision and, given that experts perform significantly better than novices (\cite{Rigoli_AAMAS_2020} and \textit{Supplementary Information, Sec A}), we also expected that the trained LSTM$_{NN, expert}$ and LSTM$_{NN, novice}$ models would be expertise specific. Finally, we expected that SHAP could be employed to identify differences in the decision-making processes of the expert and novice herders, or more specifically, how accurate LSTM$_{NN}$ models of expert or novice target selection behavior deferentially weighted task information (feature inputs) when making predictions.
\section{Results}\label{results}
Given players could choose to corral no target, predicting the target selection decisions of human herders corresponded to a 5-label prediction problem, with ID = 1, 2, 3, 4 corresponding to the actual targets and ID = 0 corresponding to no target. 

To train an LSTM$_{NN}$ using SML, we extracted time-series data from the data recordings of the expert-expert pairs and the successful novice-novice pairs that completed the dyadic herding task detailed in \cite{Rigoli_AAMAS_2020}. Only data from successful trials was employed. For each trial, state data was extracted from the time of task onset to when the players had corralled all four targets inside the containment area. 
At each time step, the target a herder was corralling was labeled manually by a research assistant, blind to the studies true purpose, using a custom coding software package (see Figure~\ref{fig:experiment}b and \textit{Methods} for more details). From the resulting labeled time-series data, training samples that included 48 input features were constructed of length $ t_i $ to $ t_f $, where $ t_f - t_i  = T_{seq}$ and the length of $T_{seq}$ corresponded to 25 time steps (i.e.,  $t_i =  t_f - 25$ time steps) or 1 second of system state evolution. The 48 state input features were the \textit{relative radial and angular distance} between herders and between the herders and each of the 4 targets, each herder's and target's \textit{radial and angular distance from the containment area}, and the \textit{radial velocity}, \textit{radial acceleration} and \textit{direction of motion} of each herder and target.

\begin{figure}[tbhp]
 	\centering
 	\includegraphics[width=\linewidth]{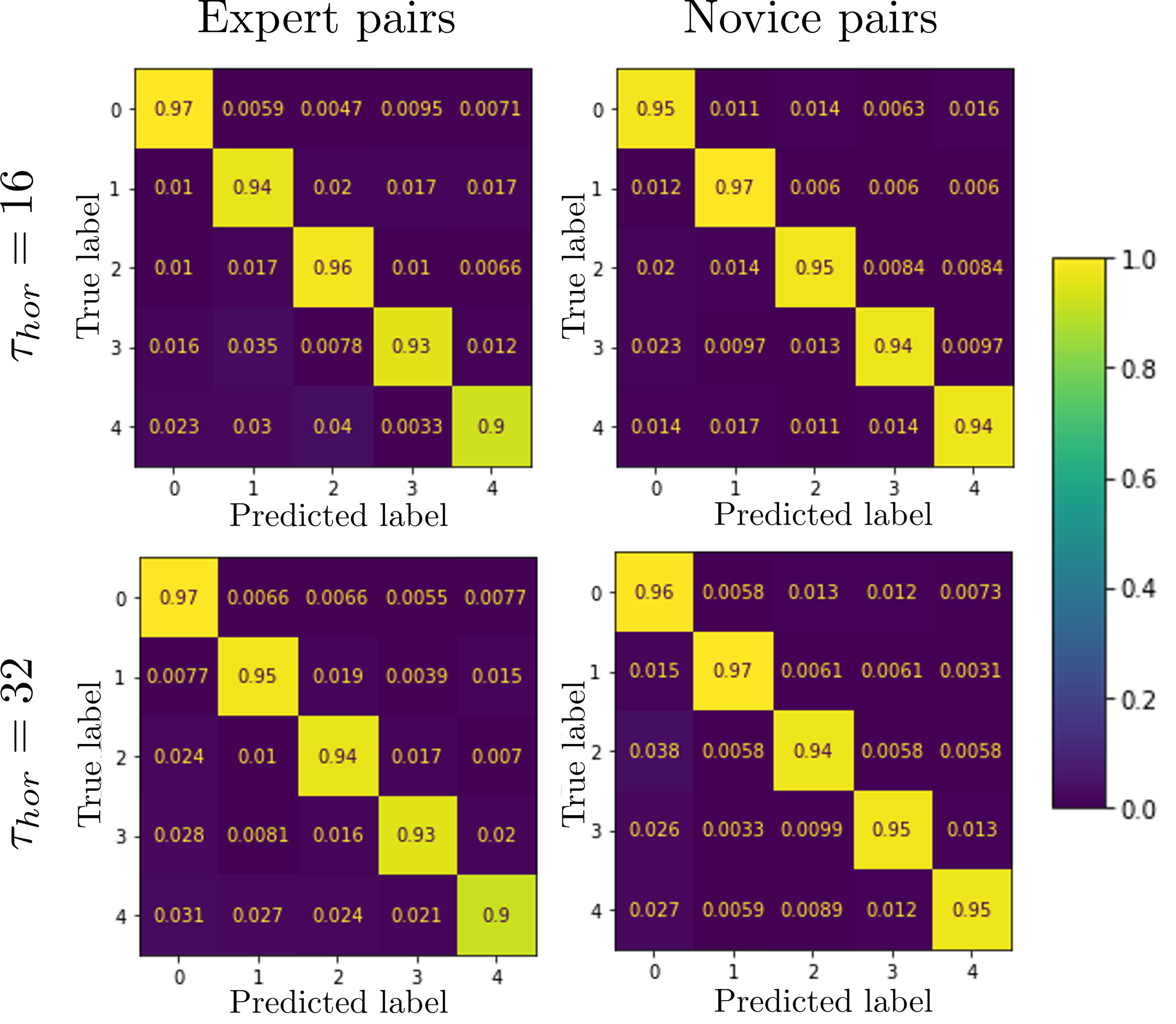}
 	\caption{Confusion matrices for the trained prediction models tested on 10 sets of $ \mathbf{N_{test}} $= 2000 samples for each expertise level and prediction horizon. Values on the diagonal indicate the portion of test samples correctly predicted.}
 	\label{fig:ConfusionMatrices}
 \end{figure}

\subsection{Predicting future target selection decisions}\label{subsec:Prediction Horizons}
LSTM$_{NN}$ models were trained to predict the next target ID a herder would corral at $ t_{f+T_{hor}}$ given a feature input sequence of length $T_{seq}$, with $ T_{hor} > 0$ time steps. Note that $ T_{hor} = 1$ corresponded to predicting the target the herder would corral at the next time-step (equivalent to 40 ms in the future) and, thus, simply entailed predicting target selection decisions already made and/or currently being enacted by a herder. Here we present the results for models trained to predict target selection decisions at two longer prediction horizons, namely, $ T_{hor} = 16$ and $32$, which corresponded to predicting the target a herder would corral 640 ms and 1280 ms in the future, respectively (for comparative purposes, we also trained models for $ T_{hor} = 1$ and $8$, see \textit{Supplementary Information, Sec. C}). 

Importantly, $ T_{hor} = 16$ and $32$ involved predicting the target selection decisions before a player's decision or behavioral intention was enacted or typically observable in the input data sequence $T_{seq}$. This was validated by calculating the average time it took players to move between targets when switching targets, with an average inter-target movement time of 556 ms for novice herders and 470 ms for expert herders (see \textit{Supplementary Information, Sec. B}).

 \begin{table}[tbhp]
     \centering
     \caption{Average performance [\%] of the expert and novel models as a function of prediction horizon, tested on 10 sets of $ \mathbf{N_{test}} $= 2000 samples.}
     \begin{tabular}{lcccc}
     	& Accuracy& Precision & Recall  & F1 score   \\
     	\midrule
     	\multicolumn{3}{l}{$ \tau_{hor} = 16$ (640 ms) prediction horizon}  &   \\
     	\midrule
     	Novice & 95.33$\pm$0.2  & 95.18$\pm$0.2  & 95.25$\pm$0.2  & 95.2$\pm$0.2   \\
     	Expert & 95.2$\pm$0.4  & 94.17$\pm$0.6 & 94.5$\pm$0.5 & 94.3$\pm$0.5  \\
     	\midrule
     	\multicolumn{3}{l}{$ \tau_{hor} = 32$ (1280 ms) prediction horizon}  &   \\
     	\midrule
     	Novice & 95.75$\pm$0.5  & 95.42$\pm$0.5  & 95.5$\pm$0.5  & 95.45$\pm$0.5   \\
     	Expert & 94.66$\pm$0.5  & 93.2$\pm$0.7 & 93.34$\pm$0.8 & 93.25$\pm$0.7  \\
     	\bottomrule
     \end{tabular}
     \label{tab:Performance}
 \end{table}

Separate LSTM$_{NN}$ were trained\footnote{Code and trained ANNs available at \href{https://github.com/FabLtt/ExplainedDecisions}{github.com/FabLtt/ExplainedDecisions}.} to predict the target selection decisions of novice and expert herders for each prediction horizon. The confusion matrices of each LSTM$_{NN}$ evaluated on ten sets of test data samples (i.e., samples not employed during training) are reported in Figure~\ref{fig:ConfusionMatrices}. Importantly, the values on the diagonal indicate that each target ID could be correctly predicted between 90\% to 97\% of the time. Indeed, independent from prediction horizon and player expertise, each LSTM$_{NN}$ predicted which target a herder would corral with an average accuracy exceeding $ 94\% $ (see Table~\ref{tab:Performance} for more prediction metrics, defined in \textit{Methods}). 

\begin{figure}
    \centering
    \begin{subfigure}[t]{0.45\textwidth}
        \centering
        \includegraphics[width=\linewidth]{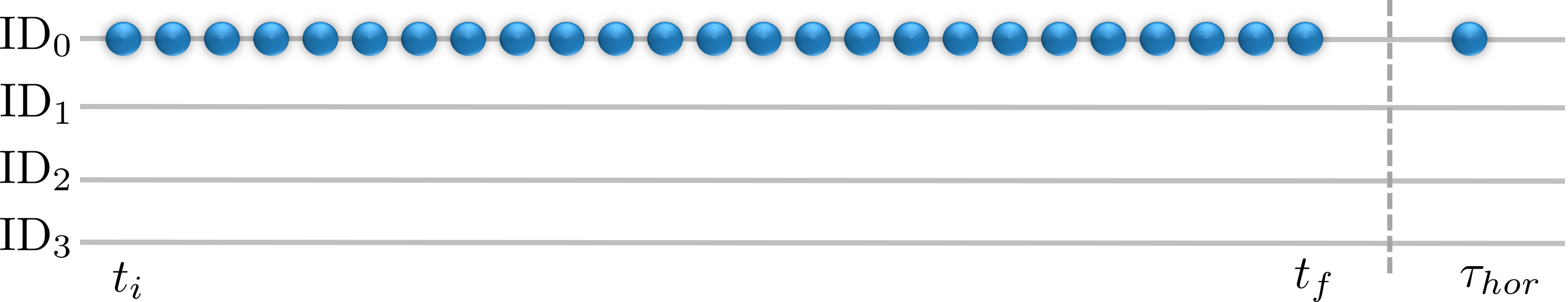} 
        \caption{Non-Transitioning, Non-Switching} \label{fig03:SamplesTypes_NTNS}
    \end{subfigure}
    \hfill
    \begin{subfigure}[t]{0.45\textwidth}
        \centering
        \includegraphics[width=\linewidth]{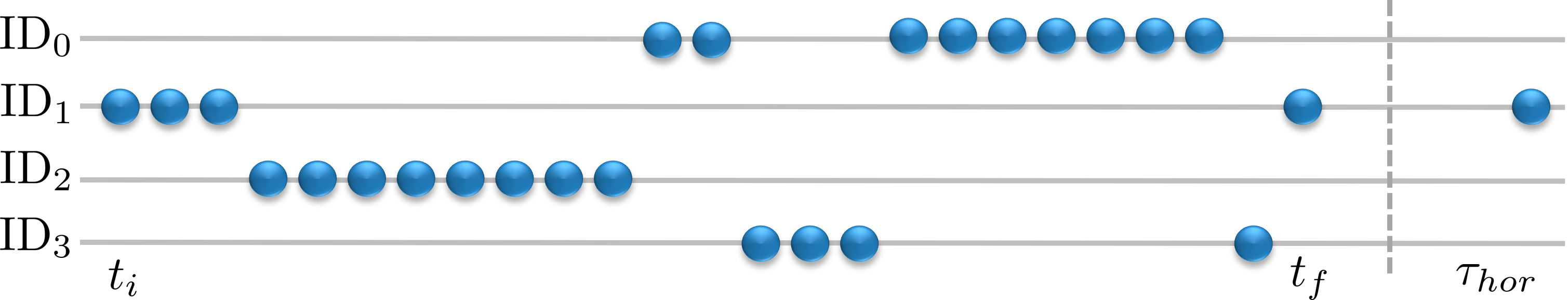} 
        \caption{Transitioning, Non-Switching}  \label{fig:SamplesTypes_TNS}
    \end{subfigure}
    
    \begin{subfigure}[t]{0.45\textwidth}
    \centering
        \includegraphics[width=\linewidth]{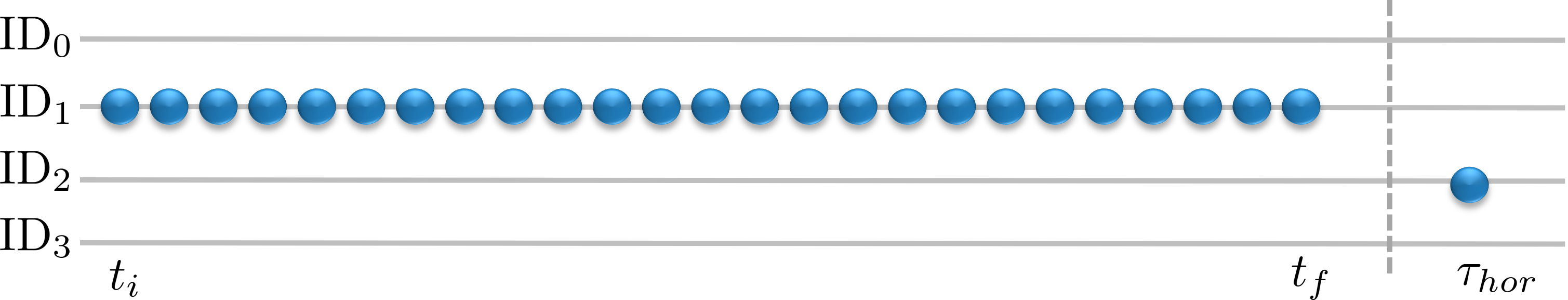} 
        \caption{Non-Transitioning, Switching} \label{fig:SamplesTypes_NTS}
    \end{subfigure}
    \hfill
    \begin{subfigure}[t]{0.45\textwidth}
    \centering
        \includegraphics[width=\linewidth]{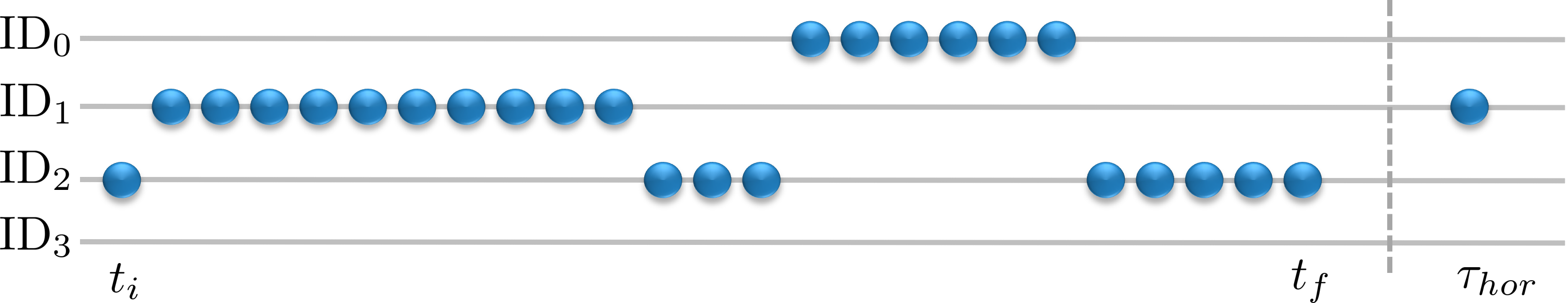} 
        \caption{Transitioning, Switching} \label{fig:SamplesTypes_TS}
    \end{subfigure}
    \caption[Examples of different sub-classes of target selection behavior.]{Illustration of the different sub-categories of target selection behavior; which also reflects the four different types of data samples used for model training and testing. Non-transitioning (a,c) and transitioning (b,d) corresponded to whether a harder corralled the same target or different targets during the input sequenced $T_{seq} = t \in [t_i,t_f] $. Non-switching (a,b) and switching (c,d) corresponded to whether a herder was corralling the same or a different target, respectively, at $T_{hor}$ and $t_f$.}
    \label{fig:SamplesTypes}
\end{figure}

\subsection{Predicting differing target selection behaviors }\label{subsec:TrainingSamples}	
Recall that the data samples used to make a target selection prediction are vector time-series of the herding system's state evolution for $ t \in [t_i,t_f] $, where $ t_f - t_i  = T_{seq}$, and the prediction outputs are chosen as the ID of the target that will be corralled at $ t_{f+T_{hor}} $ with $ T_{hor} \neq 0$. It is important to appreciate that during the time interval $ T_{seq} $, a human herder could either continuously corral the same target agent or \textit{transition} between different targets. Here we classified these as \textit{non-transitioning} and \textit{transitioning} behavioral sequences, respectively. Furthermore, at $ T_{hor} $, a herder could be corralling the same target that was being corralled at the end of $ T_{seq} $ or could \textit{switch} to a different target. Here we classified these two possibilities as \textit{non-switching} and \textit{switching} behaviors, respectively. Taken together, this defines four different sub-categories (sub-classes) of target selection behavior (or data sample type), which are illustrated in Figure~\ref{fig:SamplesTypes}.  

\begin{figure}
    \centering
    \begin{subfigure}[t]{0.4\textwidth}
        \centering
        \includegraphics[width=\linewidth]{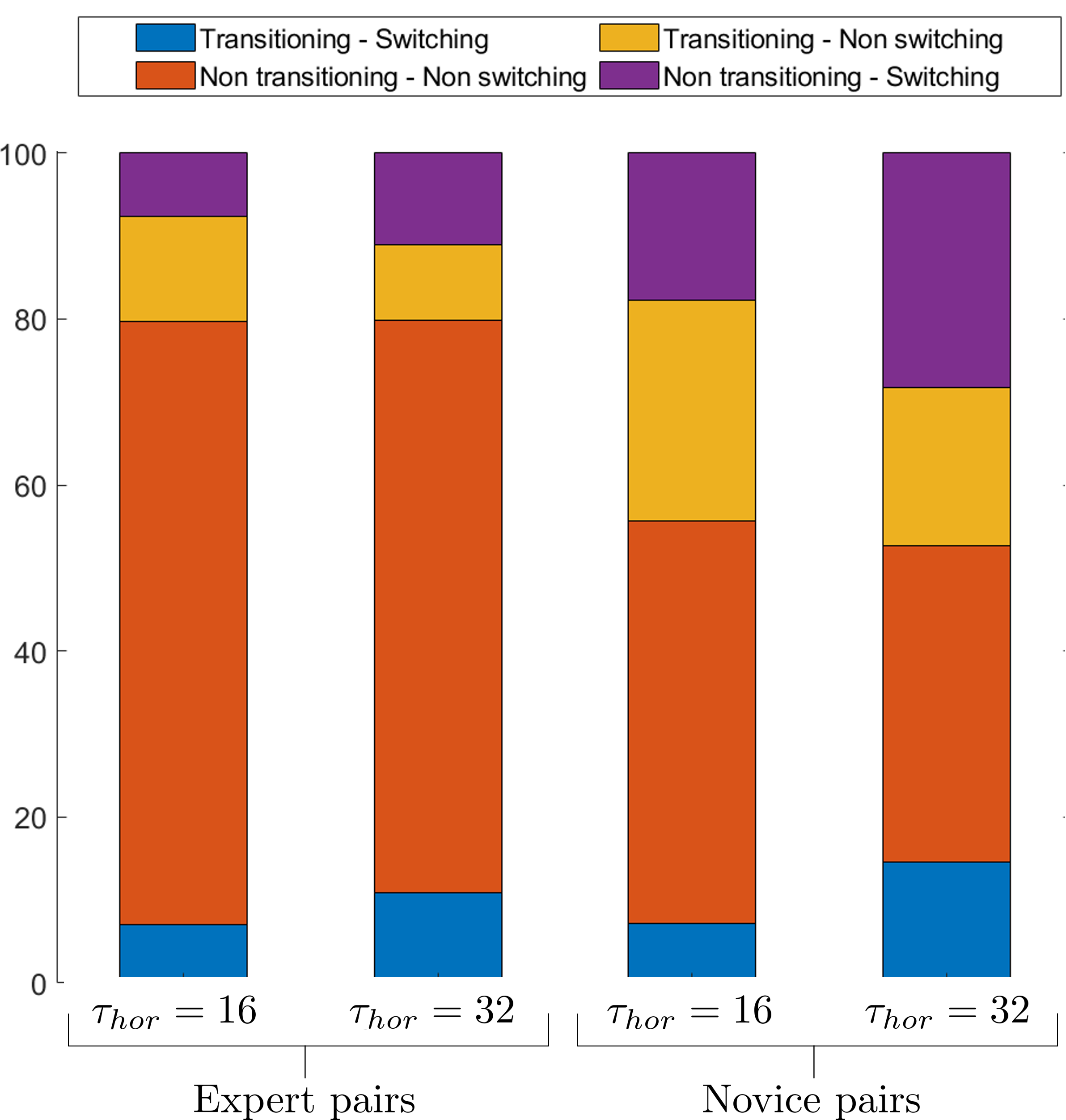} 
        \caption{} \label{fig:SamplesTypes_Distribution}
    \end{subfigure}
    \hfill
    \begin{subfigure}[t]{0.55\textwidth}
        \centering
        \includegraphics[width=\linewidth]{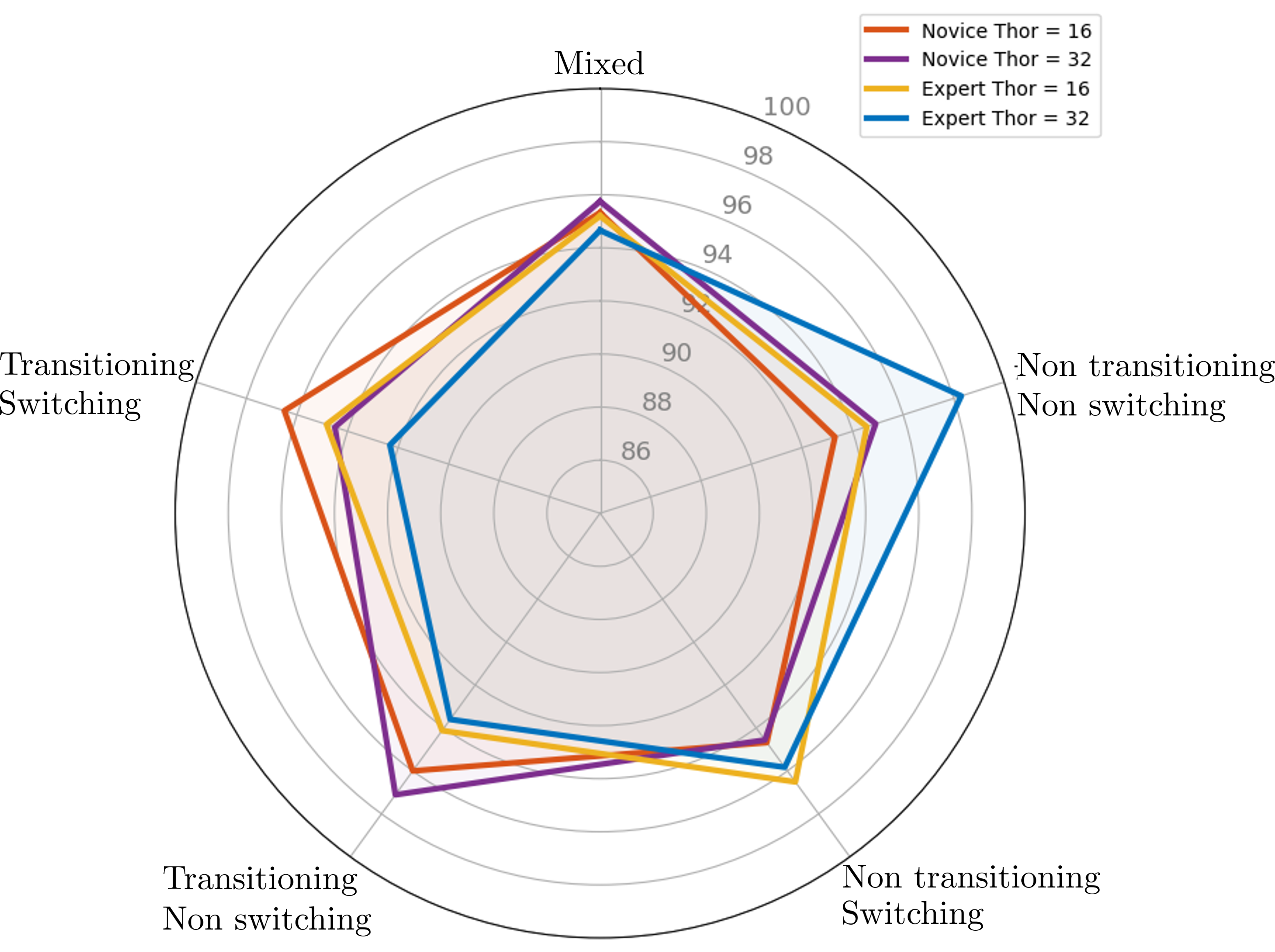} 
        \caption{} \label{fig:SpiderPlot}
    \end{subfigure}
    \caption{(a) Distribution of the different sub-categories of target selection (sample type) as a function of expertise level and prediction horizon. (b) The general (mixed) and sample specific accuracy of models trained using a uniform distribution of training samples (i.e, training set contained 25\% of each sample type) as a function of expertise level and prediction horizon. Accuracy values ranged from 92.3\% to 98.2\%, with an overall mean accuracy of 95.23\%.}
    \label{fig:SampleDistributions_SpiderPlot}
\end{figure}

The sample distribution of the different target selection behaviors observed within the expert and novice data set as a function of prediction horizon, $ T_{hor} = 16$ and $32$, are displayed in Figure~\ref{fig:SamplesTypes_Distribution}. Interestingly, the expert data contained a high proportion of non-transitioning-non-switching behavior ($69\%$), whereas the novice data-set contained a more even distribution of sample type compared to experts, particular for $T_{hor} = 32$. This indicated that experts both transitioned and switched between different targets less often than novices and were more persistent in corralling a given target compared to novices. Of more significance with regard to target selection predictions was that differences in sample type distribution could skew model accuracy due to an uneven representation of each sample type during training. That is, models trained using randomly selected training sets would exhibit lower accuracy for the underrepresented behavior types; e.g., model accuracy would be lower for transitioning-switching behaviors compared to non-transitioning-non-switching behaviors for example. This is illustrated in \textit{Supplementary Information, Sec. C}, where we show how model accuracy is sub-class dependent when models are trained using representative distributions of sample type. 

Accordingly, the LSTM$_{NN}$ models reported here were trained using training sets that contained a randomly selected, but uniform (balanced) distribution of sample type, such that each sub-class of target selection behavior was equally represented during training. Furthermore, in addition to examining overall model accuracy as a function of target ID (see Figure~\ref{fig:ConfusionMatrices} and Table~\ref{tab:Performance}), the LSTM$_{NN}$ models predicting expert and novice target selection decisions at $T_{hor} = 16$ and $32$ were also tested against $N \ge 1$ novel test sets composed of either (i) non-transitioning-non-switching, (ii) non-transitioning-switching, (iii) transitioning-non-switching, (iv) transitioning-switching samples. Note that each test set contained 2000 test samples and thus the number of test sets $N$ employed for model testing varied from 1 to 10 and was a function of how many samples were available post training (i.e., from the set of samples not employed during model training); see \textit{Methods} for more details. 

Average model accuracy for each sample type as a function of expertise level and prediction horizon is illustrated in Figure~\ref{fig:SampleDistributions_SpiderPlot}; for a detailed list of model accuracy values as a function of sample type see \textit{Supplementary Information, Sec. D}. The results revealed that the accuracy of the resultant models were essentially sample type independent. More specifically, for $T_{hor} = 16$ the accuracy for each sample type ranged between 93.3\% and 98.2\% for LSTM$_{novice}$ models and 93.6\%  and 97.11\%  for LSTM$_{expert}$ models, with an average mixed sample type accuracy of 95.33\% ($\pm$0.2) and 95.2\% ($\pm$0.4) for LSTM$_{novice}$ and LSTM$_{expert}$ models, respectively. Similarly, for $T_{hor} = 32$ the accuracy for each sample type ranged between 94.56\% and 96.51\% for LSTM$_{novice}$ models and 92.32\% and 96.5\% for LSTM$_{expert}$ models, with an average mixed sample type accuracy of 95.75\% ($\pm$0.5) and 94.66\% ($\pm$0.5) for LSTM$_{novice}$ and LSTM$_{expert}$ models, respectively.

\subsection{Specificity of expert and novice predictions}\label{subsec:Model}
\begin{figure}[tbhp]
 	\centering
 	\includegraphics[width=0.9\linewidth]{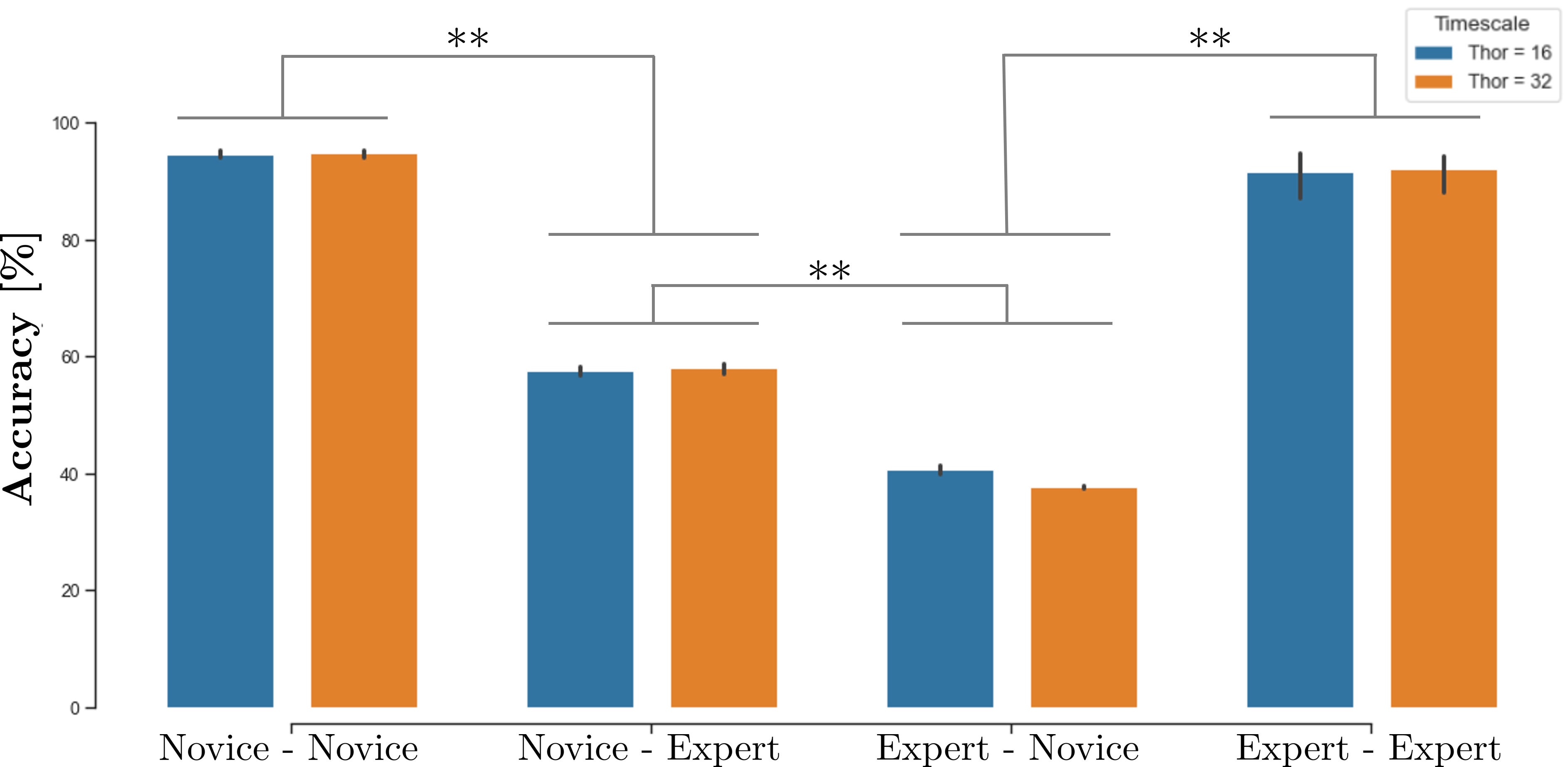}
 	\caption{Average accuracy [\%] of expert and novel trained models for 12 different tests sets of congruent (novice-novice, expert-expert) or in-congruent (novice-expert, expert-novice) $ N_{test} $= 2000 samples. $\ast \ast$ indicates a significant paired samples t-test difference of $p < .01$. Note there were no significant differences between the accuracy of LSTM$_{novice}$ and LSTM$_{expert}$ models for both prediction horizon's when tested on the same expertise level (i.e., novice-novice and expert-expert (all $p > .1$). }
 	\label{fig:CrossAccuracy}
 \end{figure}

The latter results provided clear evidence that the LSTM$_{NN}$ models could accurately predict the future target selection decisions of expert and novice herders, independent of whether the future target to be corralled was the same or different from that being corralled at $ T_{hor} \leq 0$. With regard to differentiating expert and novice performance, of equal importance was determining whether the corresponding LSTM$_{NN}$ models were expertise specific. This was tested by comparing the performance of the expert trained LSTM$_{NN}$ models attempting to predict novice target selection decisions and vice versa. As expected, when an LSTM$_{NN}$ trained on one expertise was used to predict test samples extracted from the opposite expertise model performance decreased to near or below chance levels (see Figure \ref{fig:CrossAccuracy}), confirming that the models were indeed expertise specific. More specifically, for $T_{hor} = 16$ the LSTM$_{expert}$ models predicted novice samples with an average accuracy of only 40.68\%. Similarly, the LSTM$_{novice}$ models only predicted expert samples with a 57.7\% average accuracy. For $T_{hor} = 32$, average accuracy dropped to 37.66\%  for expert-to-novice and to 58.1\% for novice-to-expert predictions.  

\subsection{Identifying differences in expert and novice target selection decisions}\label{subsec:Explain}
The significant difference in the performance of LSTM$_{NN}$ models trained and tested on the same level of expertise (see Table~\ref{tab:Performance}) compared to different levels of expertise (see Figure~\ref{fig:CrossAccuracy}) implied that the novice and expert LSTM$_{NN}$ models weighted input state variables differently. Recall that, of particular interest here was whether these differences could be uncovered using the explainable-AI technique SHAP. Accordingly, for each model we computed\footnote{Code available at \href{https://github.com/FabLtt/ExplainedDecisions}{github.com/FabLtt/ExplainedDecisions}. The computed SHAP values have been deposited in the public repository \href{https://osf.io/wgk8e/?view_only=8aec18499ed8457cb296032545963542}{https://osf.io/wgk8e/}.} SHAP values for each sample in a test set and then using the average SHAP value rank-ordered each input feature in terms of its predictive importance.

Before assessing what specific input features were weighted differently, we first computed the ordinal association of SHAP value rankings between the different LSTM$_{NN}$ models using the Kendall rank correlation coefficient (Kendall's $\tau$ \cite{mcleod2005kendall}) where $\tau=0$ corresponds to the absence of an association\footnote{Although $\tau = 0$ is the null-hypothesis (and one cannot draw conclusions from non-significant results), it does provide a robust and intuitive assessment of rank order independence.}, $\tau=1$ corresponds to perfect association (matched rankings), and $\tau=-1$ corresponds to opposite ranking orders (negative association). More specifically, we computed Kendall's $\tau$ on the SHAP rankings of the full input feature set and on the top 10 features ranked by SHAP, between the novice and expert LSTM$_{NN}$ models for each prediction horizon. Consistent with the expectation that novices and experts employed different state information when making target selection decisions, this analysis revealed little association between the novice and expert SHAP rankings for both $T_{hor} = 16$ and $32$, with an average $\tau =-.049$ ($p > .6$) for $T_{hor} = 16$ and  $\tau = .115$, ($p > .4$) for $T_{hor} = 32$ (see \textit{Supplementary Information, Sec. E} for a detailed summary of Kendall's $\tau$ values).

\begin{figure*}
     \centering
    \includegraphics[width=\linewidth]{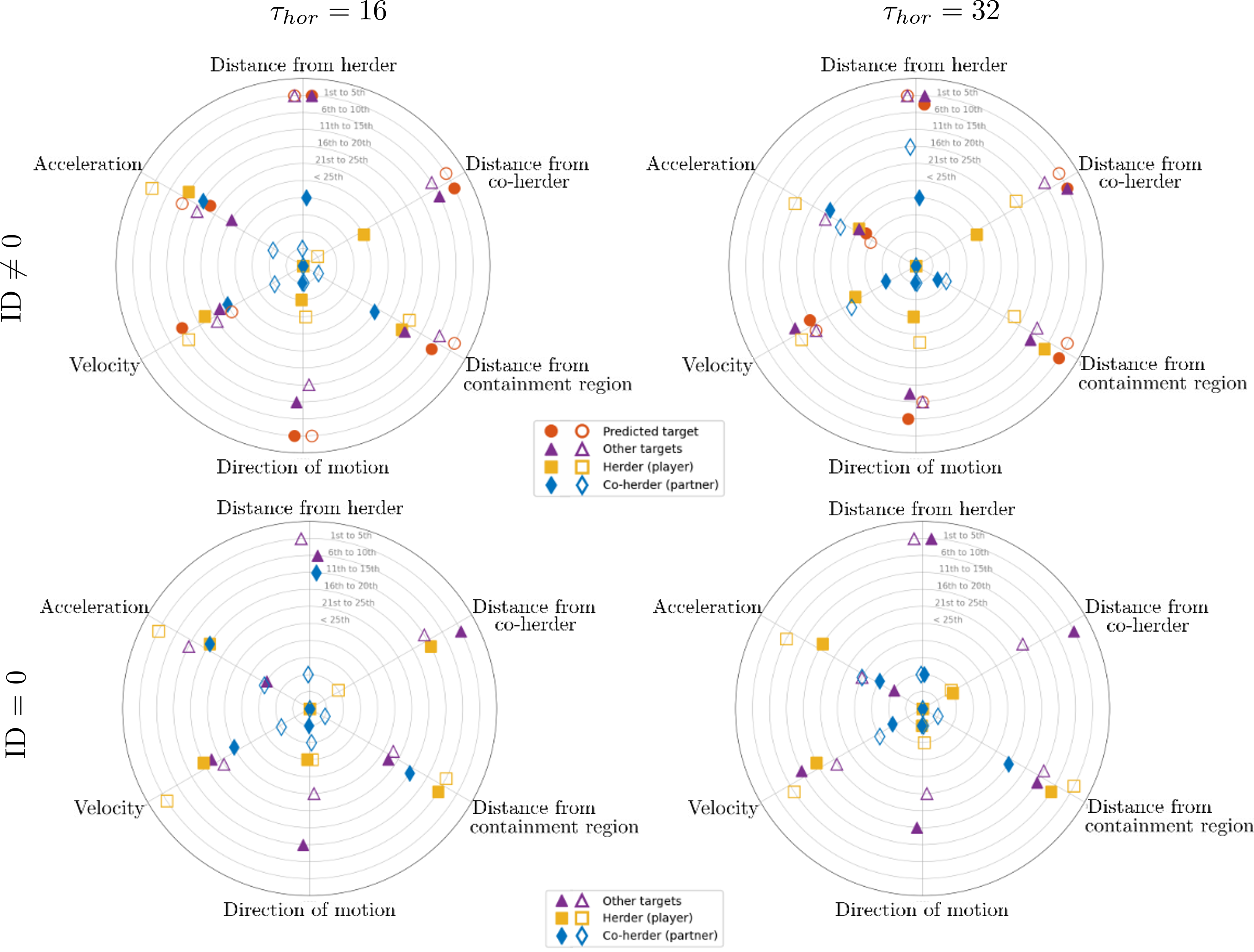}
     \caption{SHAP results for target prediction (\textit{top row}) ID $\neq 0$ (i.e., ID = 1 to 4) and (\textit{bottom row}) ID = 0, as a function of prediction horizon and expertise (novice = empty shapes, expert = filled shapes). Feature type is listed on the vertices. The radial axis represents the average rank position, such that ``1st to 5th'' represents a feature that was always or nearly always ranked as a top-five feature.}
     \label{fig:Explanation_full}
 \end{figure*}

To highlight what input features most influenced target selection predictions, Figure \ref{fig:Explanation_full} illustrates the average ranking of the different input features for both non-zero prediction outputs (i.e., ID = 1 to 4) and for ID = 0 prediction outputs, with the different input features defined as a function of input feature class (e.g., distance from herder, distance from co-herder, distance from containment area, velocity, etc.) and agent type. Note that the player that the target prediction corresponds to is referred to as the \textit{herder}, a player's partner is labeled as \textit{co-herder}, the \textit{predicted target} is the target that was predicted to be corralled by the herder at $T_{hor}$ and \textit{other-targets} corresponds to the targets that were not predicted to be corralled at $T_{hor}$ (see \textit{Supplementary Information, Sec. F} for a  detailed summary of SHAP feature values). 

Note that we split the illustration and analysis between ID $\neq 0$ and ID = 0, as there is a qualitative difference between deciding to pick a specific target to corral and choosing not to corral a target, with the latter corresponding to either a period of indecision, a period of inter-target switching, or a decision that no target needed to be collared at that time. It is also important to appreciate that there is a non-trivial difference in how one should interpret the SHAP results as a function of prediction horizon. Although accurately predicting the target selection decisions of herders further in the future provides compelling evidence of the predictive power of SML based LSTM$_{NN}$ modeling (and the potential stability or predictability of human behavior), predicting these decisions well in advance of the time that a herder actually makes the decision can provide less insight about what information the herder actually used to make their decision. As noted above, the opposite is also true, for small prediction horizons the mapping between input features and model predictions might also provide less insight about what information a herder used to make a target selection decision, as the herder is likely already enacting a made decision.

For the present work, we continued to focus on $T_{hor} = 16$ and $32$ (i.e, 640 ms and 1280 ms, respectively), as these two prediction horizons appeared to provide a good approximation of the lower and upper bounds of the timescale of herder target selection decisions. This was again motivated by the analysis of the players inter-target movement time, which as detailed above was on average less than 600 ms for both experts and novices. More specifically, for experts, 75.16\% of their inter-target movement times were less than 640 ms, with 97.38\% below 1280 ms; Similarly, for novices, 69.54\% of their inter-target movement times were less than 640ms, and 91.36\% below 1280 ms (see \textit{Supplementary Information, Sec. C}). Thus, given the fast-paced nature of the task and the assumption that herders typically decided which target to corral next just prior to action onset \cite{nalepka_herd_2017, Rigoli_AAMAS_2020}, it seemed likely that the majority of herder target selection decisions were made within this 640 to 1280 ms window.

With regard to  target ID $\neq 0$ predictions, a close inspection of Figure \ref{fig:Explanation_full}(a) revealed that the relative distance between the herder and the predicted target and the herder and the other (non-predicted) targets were consistently identified as the key input features for both expert and novice predictions. Indeed, these features nearly always ranked within the top 5 features on average, independent of prediction horizon. The distance of the predicted target from the containment region was also nearly always ranked as a top 5 feature independent of expertise and prediction horizon. Interestingly, these results are consistent with the heuristic target selection models previously developed by \cite{Nalepka_2019, Rigoli_AAMAS_2020}, in which human herders completing the same herding task explored here were assumed to pick targets that were (a) furthest from the containment region, but (b) closer to themselves than to their co-herder. Thus, the current results provide empirical support for these previous assumptions. 

Recall, that the results of the Kendall's $\tau$ analysis found very little rank-order similarity between novices and experts. Given the latter SHAP results, this implied that although expert and novice decisions appeared to be founded on similar target distance information, the specific importance ordering of target distances as a function of feature class was different across experts and novices. Indeed, a closer inspection of the top 10 feature rankings of experts and novices (see \textit{Supplementary Information, Sec. F}) revealed that for expert target selection predictions the distance of targets from the co-herder were always ranked as the most important feature, whereas for novices the distance of targets from themselves (i.e, the herder) were more often ranked as the most important feature. Although, this difference may seem subtle, it suggests that experts were more attuned to what targets better afforded corralling by their co-herder.

Further support for the latter conclusion was provided by the SHAP results for target ID = 0, where the distance of targets from a player's co-herder were consistently ranked as a top 5 feature for experts, but not for novices; see Figure~\ref{fig:Explanation_full}(b). That is, the decisions of experts to not corral a target appeared to be more evenly determined by the distance of targets from themselves (i.e, distance from herder) and their co-herder, whereas novices decisions were more heavily weight by the distance of targets from themselves. Again, this suggest that experts were more attuned to what target selection decisions were best actualized by their co-herder compared to novices. 

Consistent with the recent results of \cite{nalepka2021task}, the SHAP analysis revealed that information about the direction of motion of the predicted targets also played an important role in the target selection decisions of experts and novices. Another subtle difference between expert and novice predictions, however, was that this finding was more pronounced for experts compared to novices, with the direction of motion of the predicted target ranked as a top 10 feature on average for experts for both $T_{hor} = 16$ and $32$, as well as being more heavily weighted overall for expert target ID = 0 predictions. This implied that expert target selection decisions were more heavily influenced by whether a target was moving towards or away from the containment region. Indeed, key to task success is ensuring that targets are moving towards the containment region, irrespective of the distance from the containment area (given that targets are constrained to move within a defined ``fenced'' boundary). Thus, it is often better to choose to corral a closer target that is moving away from the containment region, than to choose a target that is further away but moving towards the containment region. 

Finally, the other difference between the SHAP results for expert and novice predictions related to the importance of herder acceleration and velocity. Indeed, these features were often ranked as a top 10 feature for novice predictions, particularly for no-target predictions when $T_{hor} = 16$, whereas they played very little role in expert predictions. Given that 29.54\% of novice inter-target movement times were greater than 640 ms, this result may be due to the novice LSTM$_{NN}$ models simply learning to map the movement information of novice herders to the periods of target ID = 0 that occur during inter-target movements. That is, rather than indicating that novices herders were influenced by their own velocity or acceleration, this may be a consequence of the slower action-decision timescales and inter-target movements of novices compared to experts. Finding that herder acceleration and velocity were less important for novice predictions when $T_{hor} = 32$ is consistent with this possibility. Thus, returning to the question of what prediction horizon best captured the decision timescale of herders, the latter result suggests that the SHAP results for the $T_{hor} = 32$ prediction horizon may better reflect the information employed by novice herders when making target selection decisions.
\section{Discussion}\label{discussion}
The current study leveraged recent advances in SML based LSTM modeling and explainable AI methods to model and understand the decision-making activity of expert and non-expert human actors performing a complex, fast-paced multiagent herding task \cite{ nalepka_herd_2017, AulettaFiore}. Results revealed that short (1 second) state information sequences ($T_{seq}$) could be used to train LSTM$_{NN}$ models to accurately predict the target selection decisions of both expert and novice players. Importantly, model predictions were made prospectively, with the majority of predictions for $T_{hor} \geq 16$ occurring before the target selection decision of herders were enacted or observable within the state input sequence. It is important to note that model effectiveness was not restricted to $T_{seq} = 1$s or $T_{hor} \geq 16$. As detailed in the \textit{Supplementary Information, Sec. C-D}, LSTM$_{NN}$ models trained using sequence lengths of .5 to 2s could accurately predict (above 95\%) target selection decision at prediction horizons ranging from 20 ms to 2.56s. Moreover, although correct predictions at $T_{hor} \geq 16$  does not provide definitive evidence that these predictions preceded a herders intent, this possibility seems likely as the action decisions made by human actors during skillful action are spontaneously tuned responses to the unfolding dynamics of a task \cite{christensen2019memory, turvey2007action, jacobs2007direct} and, for the type of perceptual-motor task investigated here, often only occur 150 to 300 ms prior to action onset \cite{welford1980reaction}. A significant implication is that the current modeling approach could be employed for the anticipatory correction of human action decisions during task training and real-time task engagement, as well as to develop more `human-like' artificial or robotic agents (\cite{AulettaChaos}) that are capable of robustly forecasting and reciprocally adjusting to the behavior of human co-actors within human-machine interaction contexts.

An interesting avenue for future research would be to explore the degree to which the current modeling approach could be employed to predict human decision-making events across a variable prediction horizon. For instance, for the current task context this would equate to predicting target switching decisions. Future research could also explore the functional relationship between prediction horizon length and accuracy as a function of the timescale of a task and its decision-making dynamics. Of interest would be whether the approach employed here can be adapted from the fast paced decision timescales that were explored here to tasks that involve much slower decision timescales (e.g., tasks where the involved actions decisions are taken over tens of seconds or minutes). 

A key finding of the current study was that the trained LSTM$_{NN}$ models were expertise specific, in that, when the expertise level of the training and test data was mismatched, prediction performance dropped to near chance levels. Consistent with decisions of skillful actors being a function of an actor’s trained attunement to the information that best specifies what action possibilities will ensure task completion \cite{jacobs2007direct, van2015information, zhao2015line}, this resulted from the expert and novice LSTM$_{NN}$ models weighting input features differently. These differences were identified using SHAP, with average SHAP feature rankings revealing that experts were more influenced by information about the state of their co-herders and were also more attuned to target direction of motion information compared to novices. Together with finding that experts transitioned between targets less often than novices, this suggests that experts were more attuned to information that better specified the collective state of the herding system, including when and what targets afforded corralling by themselves and their co-herder.

To our knowledge, no previous research has investigated the utility of explainable-AI for understanding the decision-making behavior of human actors. To date, research on explainable-AI has predominately focused on the ability of these techniques to make AI models more understandable to human users \cite{hagras2018toward, wang2019designing} and to augment or enhance the decision-making capabilities of human users \cite{alufaisan2021does}. We openly acknowledge that employing explainable-AI and SML trained LSTM$_{NN}$ to understand human decision-making is based on two fundamental assumptions: (i) that the input features employed for model training includes the informational variables employed by human actors and (ii) that the mapping between input feature weights and model predictions is isomorphic with the actual information-decision mapping that underlies human action decisions; and that these assumptions need to be validated in future work. However, the current study does provides initial evidence that explainable-AI techniques could provide a powerful tool for understanding the decision-making processes of human actors, including what information best supports optimal task performance. Indeed, the approach proposed here could be employed across a wide array of task and informational settings (i.e., visual, auditory, haptic, linguistic, etc) and, thus, the potential implications for both basic scientific research and the applied development of decision-making assessment tools could be limitless.

 \section{Method}\label{method}

 \subsection{Human herding task and data}\label{subsec:Dataset}
 Novice and expert human performance data\footnote{Raw data available at \href{https://github.com/FabLtt/ExplainedDecisions}{github.com/FabLtt/ExplainedDecisions}.The processed datasets have been deposited in the public repository \href{https://osf.io/wgk8e/?view_only=8aec18499ed8457cb296032545963542}{https://osf.io/wgk8e/}.} from the joint-action herding experiments conducted in \cite{Rigoli_AAMAS_2020} were employed for the current study. The herding task (game), developed with Unity-3D game engine (Unity Technologies LTD, CA), required pairs ($\hat N_H = 2$) of human participants (players) to control virtual herding agents to corral and contain $\hat N_T=4$ randomly moving target agents within a designated containment area positioned at the center of a game field. The task was performed on large 70" touch screen monitors (see Figure~\ref{fig:experiment}(a)), with the human participants using touch-pen stylus to control the location of motion of the herder agents. The targets' movement dynamics were defined by Brownian motion when not being influenced by a herder agent, and when influenced by a herder agent would move in a direction away from the herder agent. During task performance, the position and velocity of all herders and targets (as well as other general game states) was recorded at 50 Hz. Pairs had a maximum of 2 minutes to corral the targets into the containment area, with task success achieved if pairs could keep the targets contained within the containment area for 20 seconds. Full details of the experimental set-up and data collection process can be found in \cite{Rigoli_AAMAS_2020}

 \textbf{\textit{Novice data}} was extracted from 40 successful trials performed by 10 different novice pairs (4 successful trials from each pair). From each trial, we extracted state data from the time of task onset to when all four target agents were first contained within the specified containment area; that is, when the herders had corralled all the agents inside the containment area for the first time. The remaining trial data was disregarded as players treat the target herd as single entity after it is contained and individual target selection decisions no longer occur \cite{Nalepka_2019, Rigoli_AAMAS_2020}. Note that a human herder was considered to be a ``novice'' if they were unfamiliar with the herding task prior to the data collection session. Novices repeated the task until they had completed the 4 successful trials included in the novice data-set (with an average of 8 unsuccessful trials per pair).

 \textbf{\textit{Expert Data}} was extracted from 48 successful trials performed by 3 pairs of human players with extensive experience (completed more than 100 successful trials) performing the simulated multiagent herding task (i.e., several authors from \cite{Rigoli_AAMAS_2020}). As with the novice data, we extracted state data from the time of task onset to when all four target agents were first contained within the specified containment area.

 \subsection{State input features}\label{subsec:Features}
 From position and velocity data recorded in the original novice and expert data-sets we extracted and derived the following $ N_{sv} = 48 $ state variables:

 \begin{itemize}\label{sec:InputFeatures}

 \item[-] the radial and angular distance ($ \Delta $, $ \Psi $) between herders,
 
 \item[-] the radial and angular distance ($\Delta_{i,j}$, $\Psi_{i,j}$) of target $i$ from herder $j$ ,
 
  \item[-] the radial and angular distance of herder $j$ or target agent $i$ from the center of the containment region.
  
 \item[-] the radial velocity and acceleration of herders ($\dot r(t),\,\ddot r(t)$) and target agents ($\dot \rho(t),\,\ddot \rho(t)$),
 
 \item[-] the direction of motion of herder and target agents

 \end{itemize}


 \subsection{Target coding}\label{subsec:TargetCoding}
 A paid research assistant, naive to the study's purpose, coded (classified) what target (or not) a given herder was corralling at each point in time via an interactive data playback Unity3D\footnote{https://unity.com/, version 2018LTS} application that played back the original recorded data-set (see Figure~\ref{fig:experiment}(b)). Data playback speed could be decreased to 1/8 speed, as well as stepped frame by frame, with each target visually labeled with a fixed number (1 to 4). At each time step, the target agent a given human herder was corralling was coded by the research assistant with an integer number $ \tilde i \in [0,\hat N_T] $, with $ \tilde i = 0 $ meaning ``no target agent being corralled'' and $ \tilde i \neq 0 $ being the class ID of the target agent being corralled at that time step. 
 
\subsection{Human inter-target movement time }
To determine the time it took expert and novice herders to move from one target to the next we calculated the inter-target movement time when switching between targets ID = 1 to 4 (i.e., we ignored switch events from or to ID = 0). We calculated the time from when a human herder moved outside the region of repulsive influence of the current target and entered the region of repulsive influence of the next target. More specifically, inter-target movement time was the difference in milliseconds between the time instant at which a herder entered the repulsive radius (i.e., 0.12m around each target agent) of the current target and their relative distance was decreasing, and the time instant at which the herder left the repulsive region around the previously corralled target and their relative distance increased. In addition to the mean results report above, see \textit{Supplementary Information, Sec. B} for the distributions of inter-target movement times for expert and novice herders.

 \subsection{Training and test set data}
 All successful trial data, per level of expertise, was stacked in a common \textit{feature processed} novice  or expert data-set along with the corresponding target codes (ID 0 to 4). From the resultant, feature processed, novice and expert data-sets we randomly extracted 2 sets of $ N_{train} = 21000 $ training samples and 20 test sets of $ N_{test} = 2000 $ samples each. The first training set and 10 test sets contained transitioning/switching samples in balanced proportion (25\% each). This data set was used to train and test the models presented here. The second training set and the remaining 10 test sets contained transitioning/switching samples in the same proportion as in the entire data-set \textit{see Supplementary Information, Sec. C to E for information about the models trained using the latter unbalanced training sets}. 
 
 Here, samples refer to pairs of state feature sequences and target label codes. Sequences are composed by $ N_{seq} = 25 $ consecutive instances of the above listed $ N_{sv} $ state variables, sampled at $ dt = 0.04$ seconds, covering $ T_{seq} = 1 $ second system evolution, and labels are the ID of the agent being targeted, selected at  $ T_{hor} = 16dt $ and $ T_{hor} = 32dt $ seconds from the end of the corresponding sequence.

 In \textit{Supplementary Information, Sec. C} we also consider values of  $ T_{seq} = 0.5 $s and $ T_{seq} = 2 $ s varying the sampling time to $ dt = 0.02$ and $ dt = 0.08$ seconds respectively. As in the default case presented in the \textit{Results}, the novice and expert models trained with these different $T_{seq}$ lengths also obtain accuracy values greater than $ 95 \% $ when tested on data from the same expertise level (e.g., expert-expert) and closer to $ 50 \% $ when tested on data from  the different level of expertise (e.g., novice-expert).

 \subsection{LSTM network and Model training}
 For each combination of expertise and prediction horizon, we trained a Long-Short Term Memory (LSTM) artificial neural network with Dropout layers \cite{lstm_1997,hinton2012improving}, using Adam optimization. We used Bayesian Optimization to tune the learning rate ($ \alpha = 0.0018 $) of the Adam optimizer and the hyperparameters of the LSTM$_{NN}$ (i.e., the number of LSTM hidden layers, number of neurons in each layer, and dropout rates). The \textit{Input Layer} and the output \textit{Dense} layer of the optimized LSTM$_{NN}$ had dimensionality ($ T_{seq} $, $ N_{sv} $) and ($ T_{seq} $, $ \hat N_T + 1 $), respectively. In the center, $ 3 $ hidden LSTM layers of $ 253 $, $ 25 $ and $ 8 $ neurons were alternated with Dropout layers of equal dimensionality. The dropout rate of each LSTM layers of the novice and expert models was 0.1145. For the dropout layers between each LSTM layer, the dropout rate was 0.0145. To avoid over-fitting, training was stopped when the logarithmic loss -- that penalises false predictions -- on the validation set stopped improving; the validation set being a randomly extracted 10$ \% $ of the training set. The LSTM$_{NN}$ was built and trained using Python 3.7.1 and Tensorflow\footnote{https://www.tensorflow.org/ , version 1.15} library.

 \subsection{Model performance}
 Performance of the LSTM$_{NN}$ were validated using the following measures: \textit{Accuracy} -- the fraction of correct predictions outputs among the samples tested; \textit{Precision} -- how valid the prediction was, that is the portion of relevant outputs among the predicted ones; \textit{Recall} -- how complete the  prediction was, that is, the portion of relevant outputs that were predicted. Note that when Precision and Recall reach $ 100\% $, false positive outputs and false negative outputs are absent, respectively. Additionally, we also report the \textit{F1 score} for model prediction's, with higher values of \textit{F1 score}, the harmonic mean of Precision and Recall, expressing how precise and robust the model prediction was.

\subsection{SHapley Additive exPlanation} Given a sample, the SHAP\footnote{https://github.com/slundberg/shap, version 0.31} algorithm assigns to each input feature an importance value. This is an estimate of the contribution of each feature to the difference between the actual prediction output and the mean prediction output. We randomly selected $ \hat N_{train}^S = 200 $ samples from the training set as background set, that is, as a prior distribution of the input space. We applied SHAP \texttt{DeepExplainer} on $ N_{test} = 6000 $ of the test samples used to evaluate performance \cite{lundberg2017unified} and obtained the corresponding SHAP values for each state variable. To derive the corresponding approximate global feature importance measure (shown in Figure~\ref{fig:Explanation_full}) we averaged over the test set, for each class of prediction output (i.e., target ID).



\section*{Acknowledgement}
The authors thank Cassandra Crone for data coding, and Lilian Rigoli, Gaurav Patil, Patrick Nalepka, Elliot Saltzman, Mark Dras, Erik Reichle, Simon Hosking, Christopher Best, James Simpson and Roberto Pellungrini, for their collaborative support. Fabrizia Auletta's was supported by a Industrial and International Leverage Award from University of Bristol and an International Research Excellence Scholarship from Macquarie University. This research was also supported by an Australian Research Council Future Fellowshoip (FT180100447) awarded to Michael Richardson and Australian Department of Defence, Science and Technology group (MyIP8655 and HPRNet ID9024).

\section*{Author Contributions}
 FA, MB, RK and MR were all involved in conceptualization, formulation and assessment of study findings, as well as manuscript preparation. FA, MB and MR contributed to data modeling. KR and MR designed and developed the herding task employed and were lead investigators on the project that collected the original data. The authors have no competing interests related to the research presented here. Correspondence should be addressed to Michael J Richardson (michael.j.richardson@mq.edu.au) or Mario di Bernardo (mario.dibernardo@unina.it).

\bibliographystyle{plain}
\bibliography{Reference}
 
\appendix

\section*{Supplementary Information}

\section{Expert and Novice Herding performance}
 As in \cite{Nalepka_2019, Rigoli_AAMAS_2020}, the herding performance of players was assessed using the following five measures. (1) \textit{Gathering time}, which is the time period $t_\mathrm{g}\in[0,\, T]$, where all the passive agents are within the containment area for the first time. 
 (2) \textit{Distance traveled by the herders} $ d_\mathrm{g} $, which is the mean distance (in meters) traveled by the herders during the time interval $[0,t_\mathrm{g}]$.
(3) \textit{Herd distance from containment region} $D_\mathrm{g}$, which captures the herders ability to keep the herd close to the containment area, calculated with respect to the center of the containment area. A smaller average distance indicates better ability of the herders to keep the herd close to the containment region.
 (4) \textit{Herd spread} $ S_\mathrm{g} $, which measures the scatter of the herd within the game field during the time interval $[0,t_\mathrm{g}]$. 
 Lower values corresponds to a more cohesive herd and consequently better herding performance. The herd spread is evaluated with respect to the area of the containment region, $A_\mathrm{cr}=\pi (r^{\star})^2$, as $S_{g,\%}=S_\mathrm{g}/A_\mathrm{cr}\cdot 100$. 
 And, (5)\textit{Containment rate} $ I_{\%} $, which measures the herders' ability to relocate one or more target agents inside the containment region. It is defined as the mean in time of the percentage of agents in the containment area during the time interval $[0,t_\mathrm{g}]$. 

\begin{table}[htbp]
\centering
\noindent
\caption{Average performance of novice and expert pairs}
\begin{tabular}{lcc}
	\midrule
	& Novice pairs & Expert pairs \\
	\midrule
	$t_\mathrm{g}$ [a.u.]  	& 26.35$\pm$8.67 	& 10.65$\pm$3.43	\\
	$d_\mathrm{g}$ [a.u.]  	& 7.19$\pm$4.08 	& 4.6$\pm$1.6 	\\
	$D_\mathrm{g}$ [a.u.]  	& 0.99$\pm$0.2		& 0.35$\pm$0.22	\\
	$S_{g,\%}$ [\%] 	   	& 3.61$\pm$1.99		& 2.7$\pm$1.48		\\
	$I_{g,\%}$ [\%]  		& 18.25$\pm$7.18 	& 16.81$\pm$6.76	\\
	\midrule
\end{tabular}
\label{tab:herding_targets_Experiments}
\end{table}	

 Performance was assessed with respect to the 48 expert and 40 novice data trials employed for model training and testing. The average and SD for each measure as a function of expertise is reported in Table~\ref{tab:herding_targets_Experiments}, with experts performing better than novices with regard to all measures. More specifically, Kruskal Wallis statistical tests revealed significant differences between Novice and Expert pairs for gathering time $t_\mathrm{g}$ ($\chi^2 = 24.67$, $p<0.0001$), distance traveled  $d_\mathrm{g}$ ($\chi^2 = 5.76$, $p<0.02$) and the average distance of the herd from the containment region  $D_\mathrm{g}$ ($\chi^2 = 24.33$, $p<0.0001$).
 
 \section{Inter-target movement times}
\begin{figure}[hbtp]
	\centering
		\includegraphics[width=0.9\linewidth]{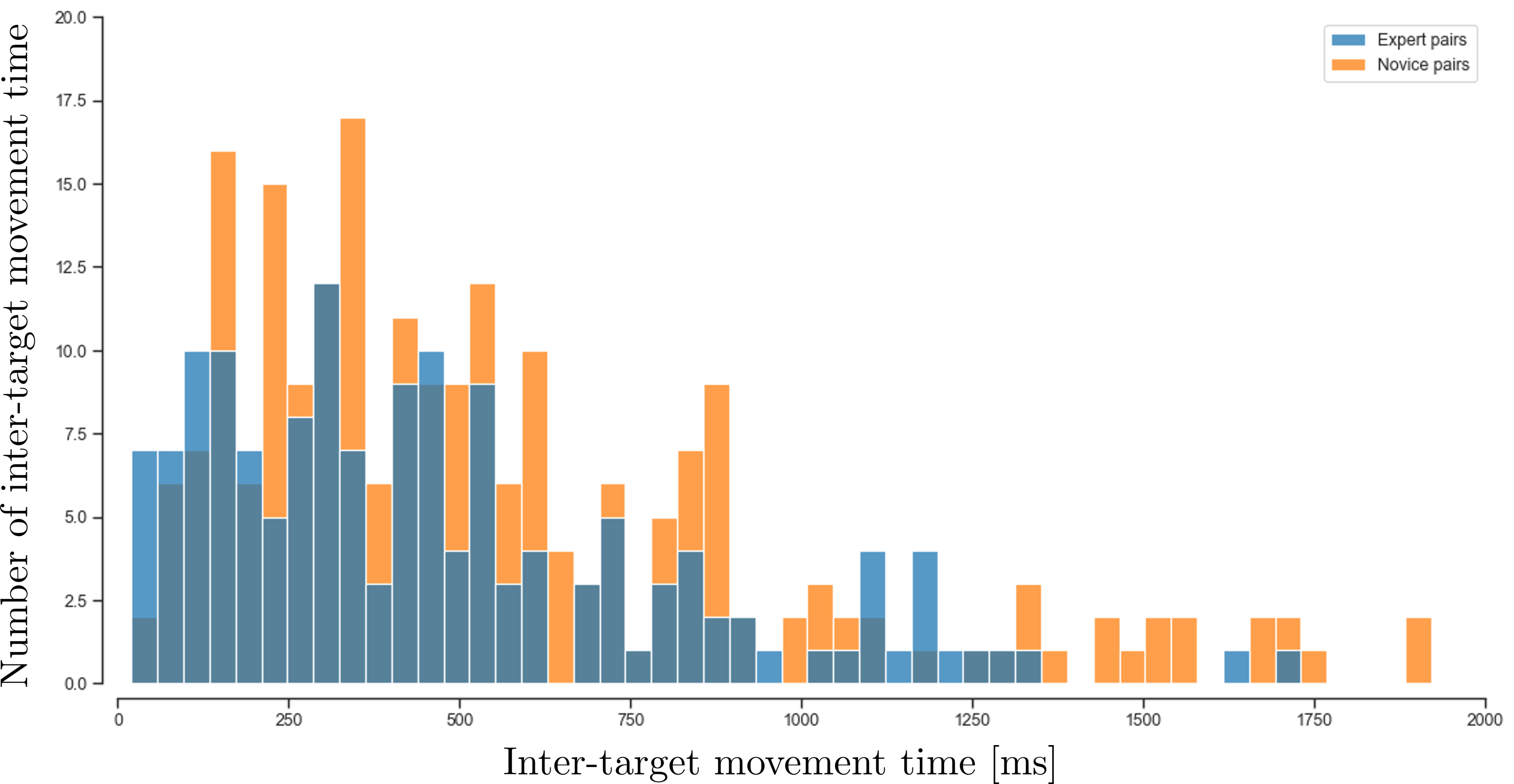}
	\caption[Distribution of inter-target movement times]{Distribution of inter-target movement times [ms] of expert (blue) and novice (orange) herders. The average inter-target movement time was 556 ms (65\% of the total inter-target movement times $ < 600 $ ms) for novices and 470 ms (72.5\% of the total inter-target movement times$ < 600 $ ms) for experts.}
	\label{fig:timeLag}
\end{figure}

For each successful trial, the inter-target movement times of experts and novice herders were determined by calculating the difference between the time a herder began influencing it's current target and the time the herder stopped influencing the previous target. Figure~\ref{fig:timeLag} reports the distribution of inter-target movement times for both expert and novice pairs. The average inter-target movement time was 556 ms for novices (with 65\% of the total inter-target movement times $\leq 600ms$) and 470 ms for experts (with 72.5\% of the total inter-target movement times $\leq600ms$).

\section{Performance of target selection models with different sequence lengths and prediction horizons}
The SML approach presented in the main article can be customized to forecast the ID of the target that will be corralled by a herder for different lengths of state input sequence, $ T_{seq} $, and different prediction horizons $ T_{hor} $. Here, $T_{seq}$ corresponded to a time-series of relevant state variables, fixed to $ N_{seq} = 25 $, such that $ T_{seq} $ is scaled by tuning the sampling time $ dt $. The output prediction is the ID of the next target to be corralled by the herder at $ T_{hor} $ in the future. 

In the main text we reported the accuracy of models trained using $T_{seq} = 1$ second of system state evolution (i.e., $ dt = 2$ or 40 ms) and prediction horizons $T_{hor} = 16dt$ and $32dt$, which corresponded to prediction horizons of 640 ms and 1280 ms, respectively. However, we also trained models for $T_{seq} = $ .5, and 2 seconds (where $ dt = 1$ and 4 time steps or 20 and 80 ms, respectively) and for $T_{hor} = 1dt$ and $8dt$. Thus, $T_{hor}$ ranging from 20 to 640 ms for $T_{seq} = .5$ seconds and from 80 ms to 2.56 seconds for $T_{seq} = 2$ seconds. 

\begin{figure}
	\centering
	\includegraphics[width=\textwidth]{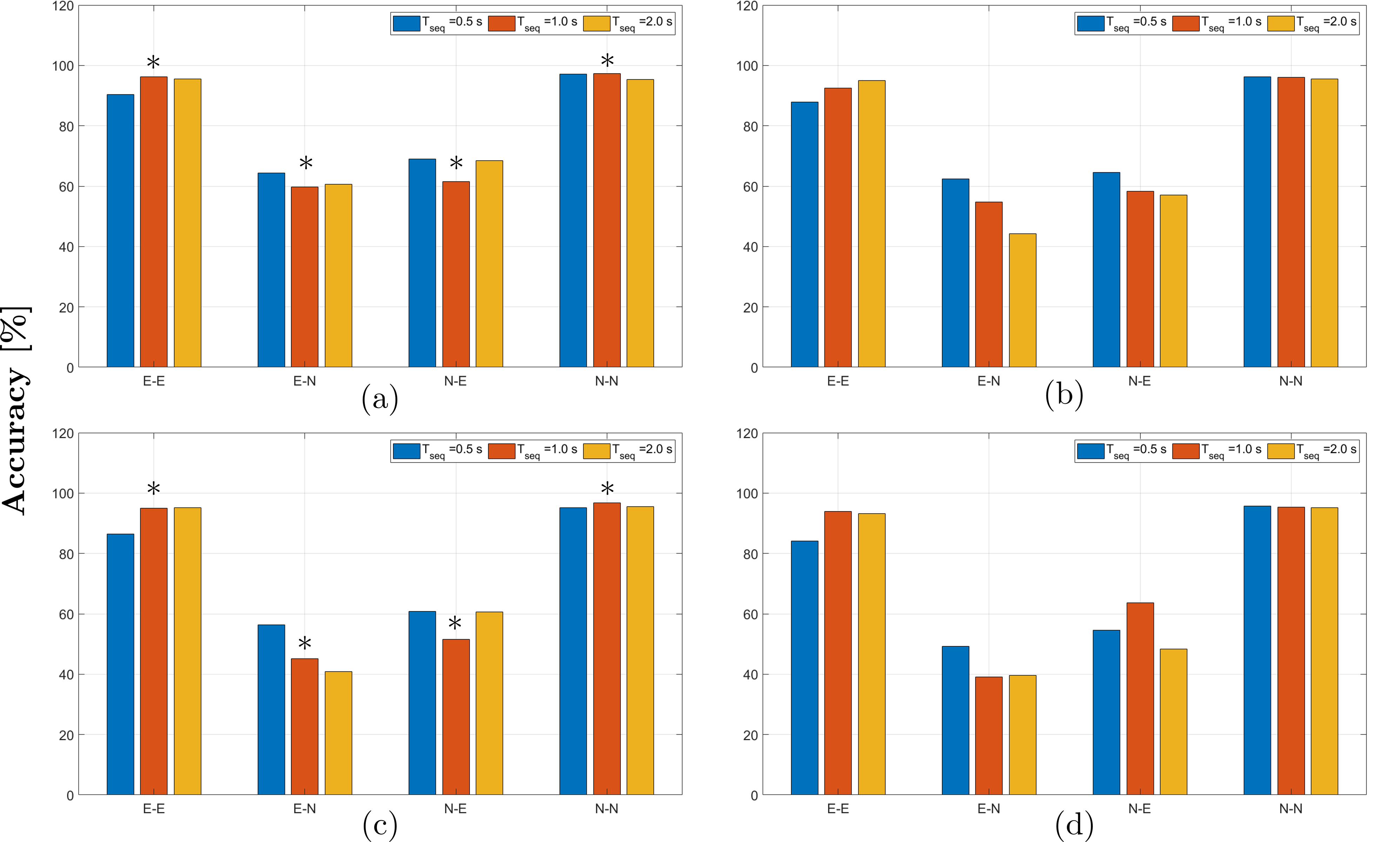}
	\caption{Overall accuracy of LSTM$_{NN}$ models trained on $ N_{test}= 2000 $ unbalanced (representative) samples for different combination of training and test pairs (E = expert; N = novice) for (a) $ T_{hor} = dt $, (b) $ T_{hor} = 8dt $, (c) $ T_{hor} = 16dt $, (d) $ T_{hor} = 32dt $. $\ast$ indicates the accuracy values for the nominal case $T_{seq} = 1~\mathrm{s}$ reported in the manuscript.}
	\label{fig:CrossAccuracy_All}
\end{figure}

The accuracy values for the different combinations of $T_{seq}$ and prediction horizon $T_{hor}$ are reported in Figure~\ref{fig:CrossAccuracy_All}. Overall, model accuracy was relative stable across the different combinations of $ T_{seq} $ and $ T_{hor}$ for both experts and novice models. Consistent with the results reported in the main text the models were also expertise specific for all combinations of $T_{seq} $ and $ T_{hor}$.

As mentioned in the main text, it is important to understand that when $ T_{hor} < 600$ ms the prediction horizon entailed predicting a target selection decision that had already been made by a herder. Thus, for $ dt = 1$ or 2 time steps, the $T_{hor} = 1dt$ and $8dt$ prediction horizons were of less interest here as the input data sequence would have included data from the enactment of the already made target selection decision (i.e., predictions were based on herder state information already specifying the made decision). They do, however, provide a benchmark measure of accuracy for the $T_{hor} = 640$ ms and 1280 ms models presented in the main text and for the $T_{seq} = 2$ second models for $T_{hor} \geq 8dt$. That is, $T_{hor} < 600$ ms predictions provide a measure of model accuracy when the target selection decision is potentially well specified in the input data (particularly for $ T_{hor} = 1dt$). The fact that the model accuracy for $T_{hor} > 600$ ms was comparable to $ T_{hor} << 600$ ms illustrates the robustness of the proposed SML-LSTM approach for predicting the target selection decisions both pre- and post-enactment.

\section{Performance of target selection models with different type of samples}
As detailed in the main text, during the time interval $ T_{seq} $, a herder could either continuously corral the same target agent or transition between different targets. These were classified as ``non-transitioning'' and ``transitioning'' behavioral sequences, respectively. Similarly, at $ T_{hor} $, a herder could be corralling the same target agent that was being corralled at the end of $ T_{seq} $ or ``switch'' to different target agent. These were classified as ``non-switching'' and ``swithching'' behavioral sequences, respectively. The resultant four data sample types are illustrated in Figure 3 in the main text.

\begin{figure}
	\centering
	\includegraphics[width=\textwidth]{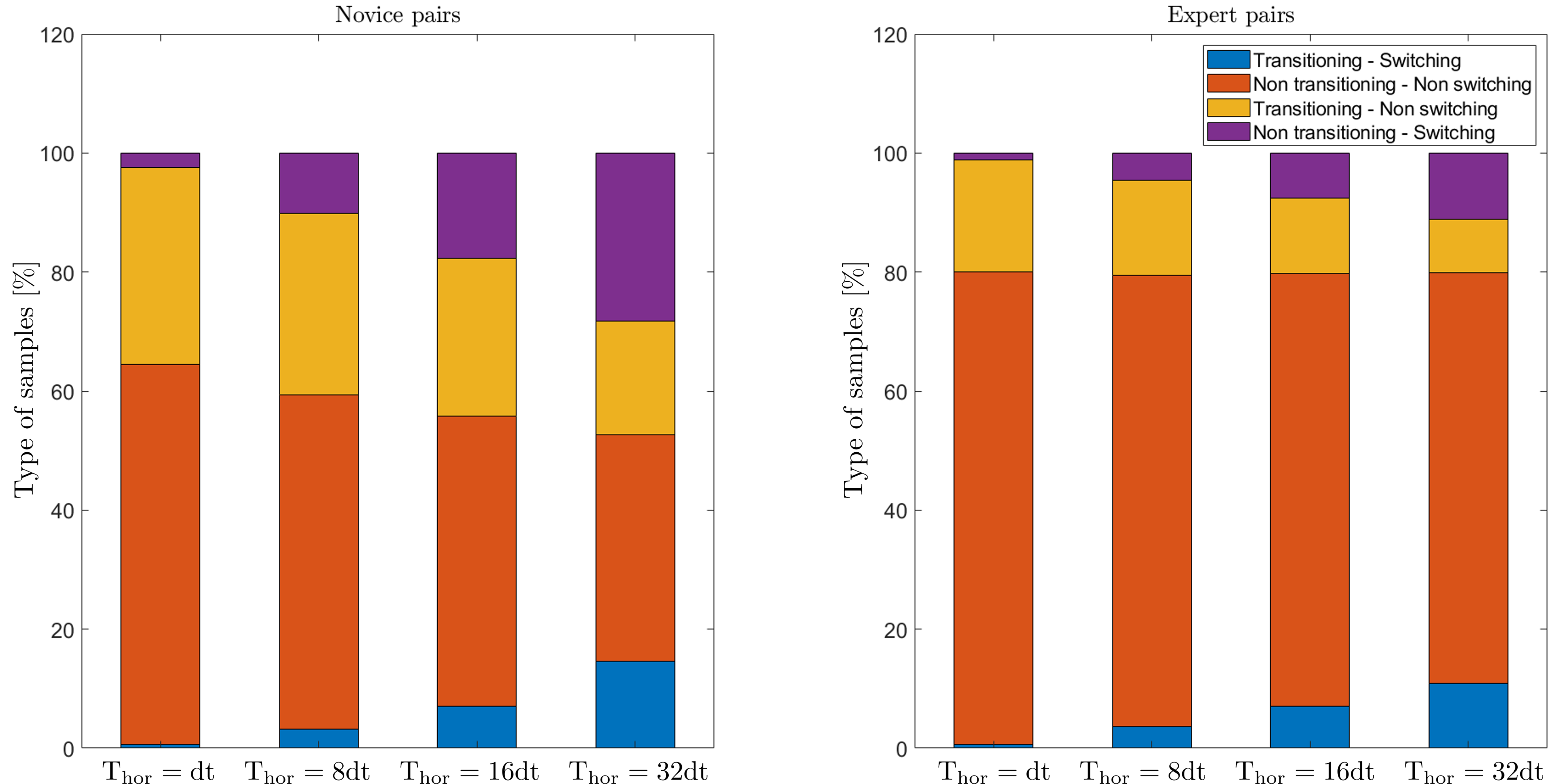}
	\caption{Percentage different type of samples in the training set for different prediction horizon and $ T_{seq} = 1 $s.}
	\label{fig:SamplesTypes_trainingsets}
\end{figure}

Importantly, the number of "switching" samples within the data set used for model training and testing was dependent on $T_{hor}$. Indeed, both novice and expert sample data contained less than 2\% transitioning-switching and less then 3\% non-transitioning-switching samples when $T_{hor} = 1$, and less than 5\% transitioning-switching and less than 7\% non-transitioning-switching samples when $T_{hor} = 8$. The different distributions of sample type as a function of $T_{hor}$ is illustrated in Figure~\ref{fig:SamplesTypes_trainingsets}.

\begin{figure}
	\centering
	\includegraphics[width=\textwidth]{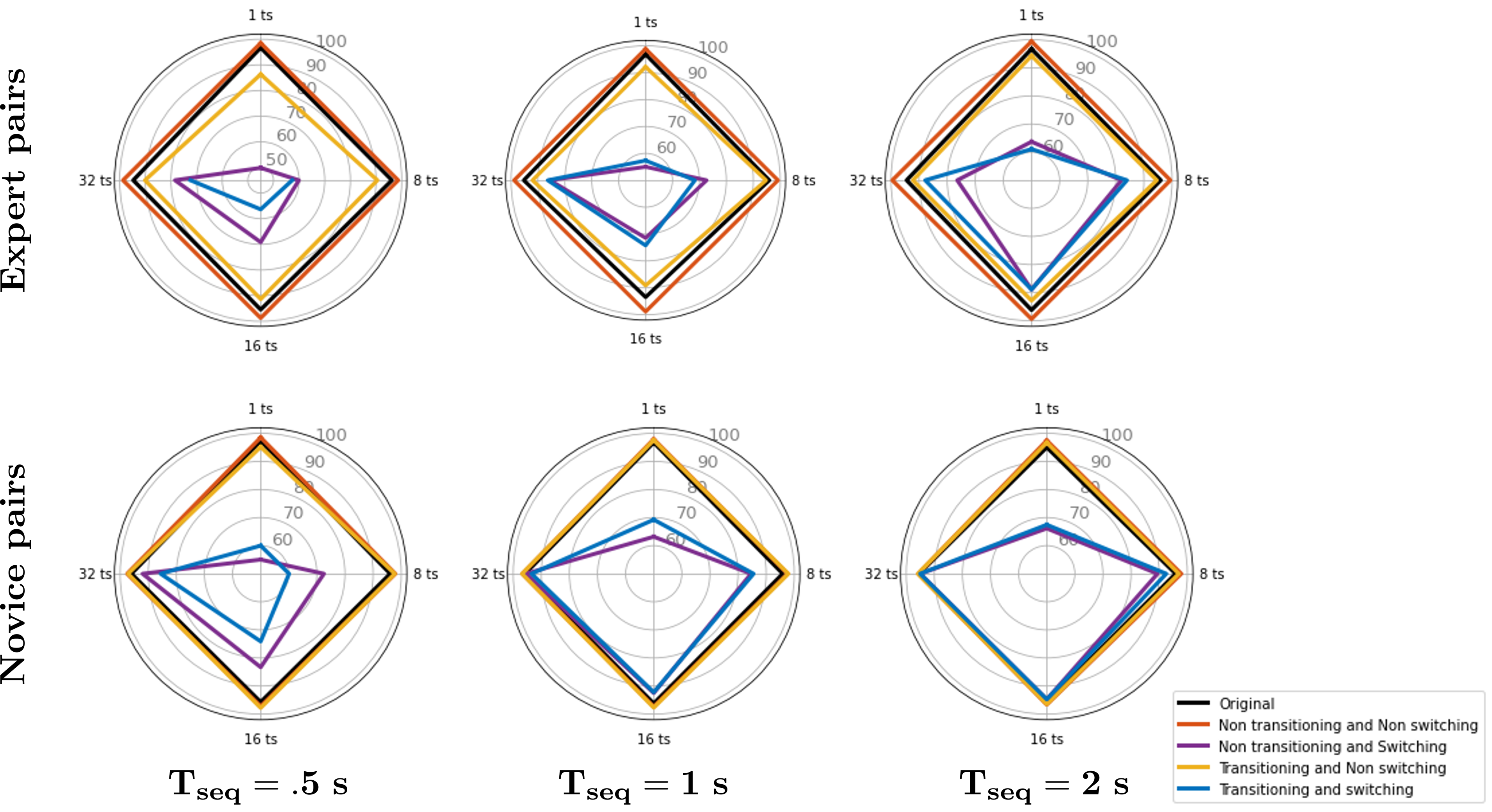}
	\caption{Accuracy of models trained using unbalanced (representative) training sets, as as function of sequence length, prediction horizon, expertise and sample type. Vertexes are the accuracy of the trained models for each decision time interval $ T_{seq} $ and horizon $ T_{hor} $. Accuracy for each model is scored on $ \mathbf{N_{test}} $= 2000 samples of the corresponding  sample type or the 'original" 2000  mixed samples).}
	\label{fig:Accuracy_All}
\end{figure}

Figure~\ref{fig:Accuracy_All} details the performance of the trained LSTM$_{NN}$ models on $N_{test} = 2000$ test samples randomly extracted from the different sample types. That is, in contrast to the models presented in the main text, the accuracy values reported in Figure~\ref{fig:Accuracy_All} reflect the accuracy of LSTM$_{NN}$ models trained on an unbalanced (representative) set of training samples. Not surprisingly, the accuracy of the models for a specific type of sample was dependent on the proportion of samples with the data set, with the accuracy for switching samples greatly reduced for $ T_{hor} = 1dt, 8dt$ and $16dt$ (particularly when $T_{seq} = .5$ seconds; i.e., when  $ dt = 1$ or 20 ms), because of the reduced number of switching samples in the training set. Indeed, there is a direct correspondence between the accuracy reported in  Figure~\ref{fig:Accuracy_All} and the proportion of a given sample type illustrated in Figure~\ref{fig:SamplesTypes_trainingsets}; also see Table~\ref{tab:PerformanceTypeSamples}. It is for this reason that the models reported in the main text were trained on balanced (uniform) data sets (i.e., training sets that included an equal number of each sample type) in order to ensure that model accuracy was sample type independent. Note, however, that balanced training was never possible for $ T_{hor} = 1dt$, as there are never enough switching samples, even when $T_{seq} = 2$ seconds. 

For comparative purposes, the accuracy of models trained using unbalanced (representative) and balanced (uniform) training sets for $ T_{hor} = 16$ or 640 ms and $ T_{hor} = 32$ or 1280 ms, when $T_{seq} = 1$ second, are detailed in Table~\ref{tab:PerformanceTypeSamples} and Table~\ref{tab:PerformanceTypeSamples_balanced}, respectively. The accuracy values in Table~\ref{tab:PerformanceTypeSamples_balanced} correspond to those reported in the main text.

\begin{table}
	\centering
	\caption[Average prediction performance of multi-label predictor for $ T_{seq} = 1 $ s, per type of samples]{Sample type performance [\% accuracy] for $T_{hor} = 16$ and $T_{hor} = 32$ models, when $T_{seq} = 1s$, trained using a representative (unbalanced) distribution of sample type. The \% of each sample type with the data set is also reported. Models were tested on a corresponding sets of $ N_{test} $= 2000 samples.}
	\resizebox{\linewidth}{!}{%
	\begin{tabular}{lcccccccccccccccccc}
		\toprule
		& \multicolumn{4}{c}{Non transitioning} & \multicolumn{4}{c}{Transitioning}  & \multicolumn{2}{c}{Mixed}  \\
		\midrule
		& \multicolumn{2}{c}{Non switching} & \multicolumn{2}{c}{Switching} & \multicolumn{2}{c}{Non switching} & \multicolumn{2}{c}{Switching} & \multicolumn{2}{c}{} \\
		\midrule
		& Accuracy & $ \% $ samples & Accuracy & $ \% $ samples & Accuracy & $ \% $ samples & Accuracy & $ \% $ samples & Accuracy & $ \% $ samples \\
		\midrule
		\multicolumn{11}{l}{$ \tau_{hor} = 16$ (640 ms) prediction horizon}   \\
		\midrule
		Novice & 97.35\%	& 48.61\%	& 92.48\% 	& 17.69\%	& 97.69\%	& 26.59\%	& 92.22\%	& 7.12\% 	& 96.21\% 	& 100\% 	\\
		Expert & 98.86\%    & 72.67\%   & 71.45\%   & 7.63\%    & 89.14\%   & 12.62\%   & 74.24\%   & 7.08\%    & 93.48\%   & 100\%    \\
		\midrule
		\multicolumn{11}{l}{$ \tau_{hor} = 32$ (1280 ms) prediction horizon}  \\
		\midrule
		Novice & 96.72\%	& 38.14\%	& 95.48\% 	& 28.27\%	& 96.84\%	& 19.02\%	& 93.48\%	& 14.57\% 	& 95.78\% 	& 100\% 	\\
		Expert & 98.86\%    & 69.06\%   & 85.48\%   & 11.11\%    & 92.03\%   & 9\%   & 86.37\%   & 10.84\%    & 95.27\%   & 100\%    \\
		\bottomrule
	\end{tabular}	}
	\label{tab:PerformanceTypeSamples}
\end{table}

\begin{table}
	\centering
	\caption[Average prediction performance per balanced type of samples]{Sample type performance [\% accuracy] for $T_{hor} = 16$ and $T_{hor} = 32$ models, when $T_{seq} = 1s$, trained using a uniform (balanced) distribution of sample type (i.e, training set contained 25\% of each sample type). Models were tested on a corresponding sets of $ N_{test} $= 2000 samples.}
	\resizebox{\linewidth}{!}{%
	\begin{tabular}{lccccc}
		\toprule
		& \multicolumn{2}{c}{Non transitioning} & \multicolumn{2}{c}{Transitioning}  & \multicolumn{1}{c}{Mixed}  \\
		\midrule
		& {Non-switching} & {  Switching  } & {Non-switching} & {  Switching  } \\
		\midrule
		\multicolumn{6}{l}{$ \tau_{hor} = 16$ (640 ms) prediction horizon}  \\
		\midrule
		Novice & 93.3 & 94.91 & 94.56 & 98.2 & 95.33$\pm$0.2 \\
		Expert & 96$\pm$0.3 & 97.11$\pm$0.3 & 94.12$\pm$0.9 & 93.6$\pm$0.6 & 95.2$\pm$0.4  \\
		\midrule
		\multicolumn{6}{l}{$ \tau_{hor} = 32$ (1280 ms) prediction horizon}  \\
		\midrule
		Novice & 94.68$\pm$0.8 & 94.56$\pm$0.4 & 96.51$\pm$0.6 & 95.83$\pm$0.3 & 95.75$\pm$0.5 \\
		Expert & 96.5$\pm$0.5 & 94.5$\pm$0.5 & 94.83$\pm$0.2 & 92.32$\pm$0.73 & 94.66$\pm$0.5  \\
		\bottomrule
	\end{tabular}	}	
	\label{tab:PerformanceTypeSamples_balanced}
\end{table}

\section{Kendall rank correlation coefficients}
The ordinal association of SHAP value rankings (the first top 10 reported in  Tables~\ref{tab:SHAP_novice_1sec_balanced}-\ref{tab:SHAP_expert_1sec_balanced}) was computed using the Kendall rank correlation coefficient (Kendall's $\tau$) for subgroups of N top ranked input features. 
Table~\ref{tab:KendallTau_noviceVsexpert_balanced} includes the Kendall coefficients, and associated p-values, as a function of expertise. Table~\ref{tab:KendallTau_ShortVsLong_balanced} includes the Kendall coefficients as a function of $T_{hor} = 16$ and  $T_{hor} = 32$ prediction horizon for each level expertise.

\begin{figure}[hbtp]
	\centering
		\includegraphics[width=\linewidth]{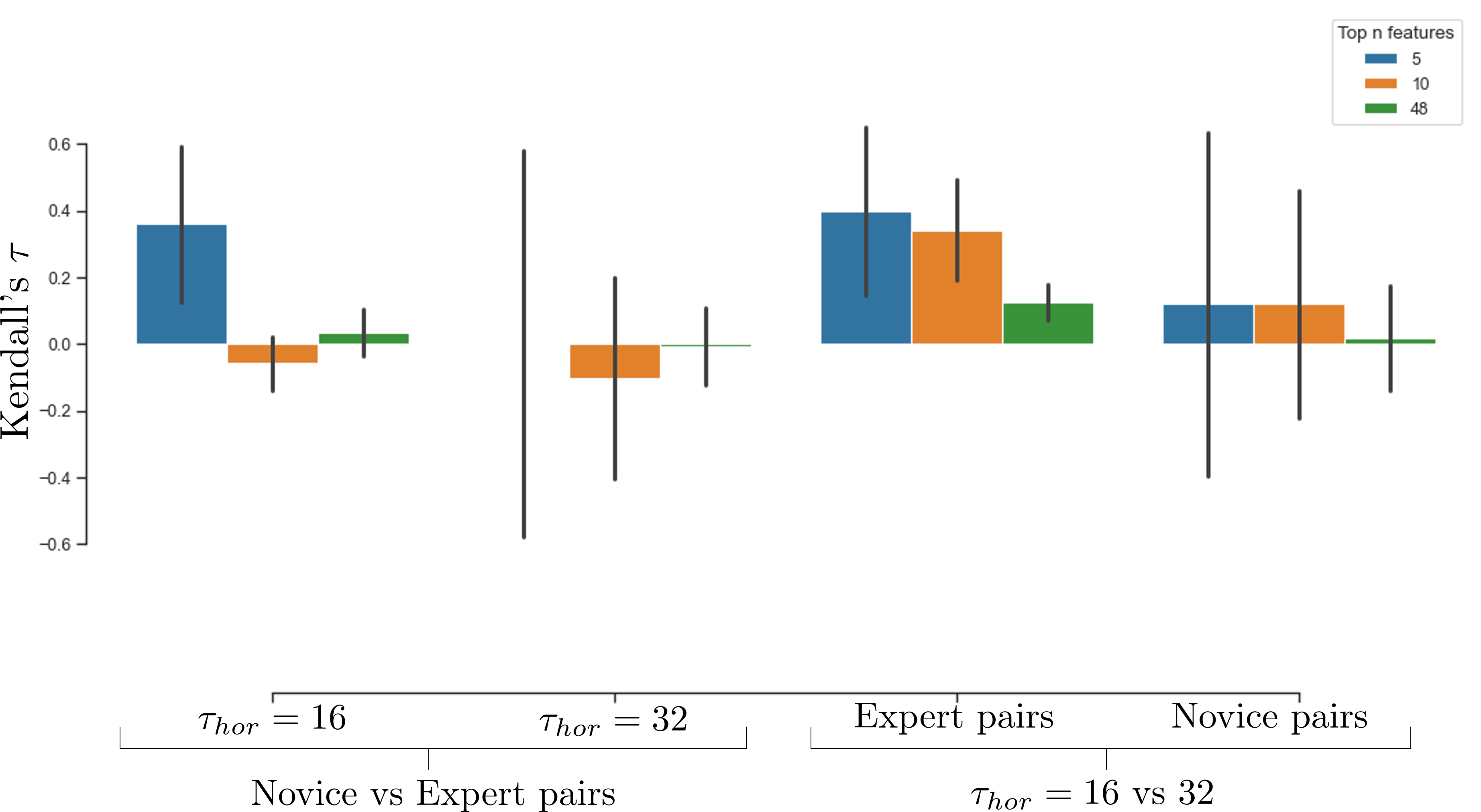}   
		\caption[Kendall's $ \tau $ values of top ranked input features]{Kendall's tau values for subgroups of top ranked input features averaged over labels for models trained on balanced data sets.} 
		\label{fig:KendallTau_bp}
\end{figure}

\section{SHAP value tables for each feature for the target selection models}
A detailed summary of SHAP feature values for each LSTM$_{NN}$ model, prediction horizon and target ID is provided in Tables~\ref{tab:SHAP_novice_1sec_balanced}-\ref{tab:SHAP_expert_1sec_balanced}.

\begin{table}
	\centering
	\caption{Kendall $\tau$'s values and corresponding p-values between novice and expert SHAP ranking for both $T_{hor} = 16$  and $T_{hor} = 32$ for models trained on balanced samples.}
	\resizebox{\linewidth}{!}{%
	\begin{tabular}{|l|ll|ll|ll|}
		\hline
		{} & \multicolumn{6}{c|}{$T_{hor} = 16$ (640 ms)}\\
		\hline
		{} & \multicolumn{2}{c|}{All features} & \multicolumn{2}{c|}{Top 10} & \multicolumn{2}{c|}{Top 5}  \\
		\hline
		{} &  Kendall tau &  p-value &  Kendall tau &  p-value &  Kendall tau &  p-value  \\
		\hline
Label 0 &                 ~0.147 &              0.140 &           ~0.022 &        1 &                0.6 &       0.233  \\
Label 1 &                 -0.034 &              0.735 &           ~0.022 &        1 &                0.6 &       0.233  \\
Label 2 &                 ~0.092 &              0.355 &           -0.155 &        0.6 &              0.4 &       0.483  \\
Label 3 &                 -0.025 &              0.803 &           -0.155 &        0.6 &              0.2 &       0.817  \\
Label 4 &                 -0.009 &              0.929 &           -0.022 &        1 &                0.0 &       1  \\
		\hline
		{} & \multicolumn{6}{c|}{$T_{hor} = 32$ (1280 ms)}\\
		\hline
		{} & \multicolumn{2}{c|}{All features} & \multicolumn{2}{c|}{Top 10} & \multicolumn{2}{c|}{Top 5}  \\
		\hline
		{} &  Kendall tau &  p-value &  Kendall tau &  p-value &  Kendall tau &  p-value  \\
		\hline
Label 0 &                 -0.161 &              0.106 &           -0.6 &        0.017 &               -1 &         0.017  \\
Label 1 &                 ~0.188 &              0.059 &           ~0.2 &        0.484 &               ~0.8 &       0.083  \\
Label 2 &                 -0.046 &              0.644 &           -0.067 &      0.862 &               ~0.0 &       1 \\
Label 3 &                 -0.051 &              0.606 &           -0.244 &      0.381 &               ~0.0 &       1  \\
Label 4 &                 ~0.041 &              0.683 &            0.2 &        0.484 &               ~0.2 &       0.817  \\
		\hline
	\end{tabular}
	}
\label{tab:KendallTau_noviceVsexpert_balanced}
\end{table}

\begin{table}
\centering
\caption{Kendall $\tau$'s values and corresponding p-values between $T_{hor} = 16$ and $T_{hor} = 32$ prediction horizons SHAP ranking for both novice and expert models trained on balanced samples.}
\resizebox{\linewidth}{!}{%
\begin{tabular}{|l|ll|ll|ll|}
	\hline
	{} & \multicolumn{6}{c|}{expert models}\\
	\hline
	{} & \multicolumn{2}{c|}{All features} & \multicolumn{2}{c|}{Top 10} & \multicolumn{2}{c|}{Top 5}  \\
	\hline
	{} &  Kendall tau &  p-value &  Kendall tau &  p-value &  Kendall tau &  p-value  \\
	\hline
Label 0 &                  0.066 &              0.511 &            0.334 &        0.216 &              0.4 &       0.484  \\
Label 1 &                  0.128 &              0.201 &            0.467 &        0.072 &              0.4 &       0.484  \\
Label 2 &                  0.126 &              0.207 &            0.556 &        0.029 &              0.4 &       0.484 \\
Label 3 &                  0.079 &              0.424 &            0.2   &        0.484 &              0 &         1 \\
Label 4 &                  0.219 &              0.027 &            0.156 &        0.601 &              0.8 &       0.084  \\
	\hline
	{} & \multicolumn{6}{c|}{novice models}\\
	\hline
	{} & \multicolumn{2}{c|}{All features} & \multicolumn{2}{c|}{Top 10} & \multicolumn{2}{c|}{Top 5}  \\
	\hline
	{} &  Kendall tau &  p-value &  Kendall tau &  p-value &  Kendall tau &  p-value  \\
	\hline
Label 0 &                 ~0.002 &              0.986 &           ~0.067 &        0.862 &               ~0 &     1  \\
Label 1 &                 ~0.25 &               0.012 &           ~0.644 &        0.009 &               ~1 &     0.017  \\
Label 2 &                 -0.019 &              0.845 &           ~0.111 &        0.727 &               ~0 &     1  \\
Label 3 &                 ~0.089 &              0.374 &           ~0.2 &          0.484 &               ~0.2 &   0.817  \\
Label 4 &                 -0.234 &              0.019 &           -0.422 &        0.108 &               -0.6 &   0.233  \\
	\hline
\end{tabular}
}
\label{tab:KendallTau_ShortVsLong_balanced}
\end{table}

\begin{table}
	\centering
	\caption{Top 10 ranked features and corresponding SHAP values for each class predicted by the model trained on \textit{novice} pairs with a sequence $T_{seq} = 1 \mathrm{s}$ and model trained on balanced data.}
	\resizebox{\linewidth}{!}{%
	\begin{tabular}{|l|ll|ll|ll|ll|ll|}
			\hline
		{} & \multicolumn{10}{c|}{$T_{hor} = 16$ (640 ms) prediction horizon}\\
		\hline
		{} & \multicolumn{2}{l|}{Label 0} & \multicolumn{2}{l|}{Label 1} & \multicolumn{2}{l|}{Label 2} & \multicolumn{2}{l|}{Label 3} & \multicolumn{2}{l|}{Label 4}\\
		\hline     
		{} & Features & SHAP  &   Features & SHAP  &  Features & SHAP  &  Features & SHAP  &  Features & SHAP  \\
		{} &  &  values &    &  values &   &  values &   &  values &   &  values \\
		\hline
	1  &    herd.1 accel. &  0.0305 &    herd.2 targ.1 dist. &  0.02450 &     herd.1  targ.2 dist. &      0.0248 &     herd.1  targ.3 dist. &      0.0234 &    herd.2 targ.4 dist. &      0.0313 \\
    2  &    herd.velocity &      0.0272 &     herd.1  targ.1 dist. &      0.0232 &    herd.2 targ.2 dist. &      0.0230 &    herd.2 targ.3 dist. &      0.0228 &     herd.1  targ.4 dist. &      0.0284 \\
    3  &    herd.1  targ.4 dist. &      0.0249 &     herd.1  targ.4 dist. &    0.0154 &  targ.2 goal dist. &        0.0201 &  targ.3 goal dist. &      0.01739 &          herd.1 accel. &      0.0183 \\
    4  &    herd.1  targ.1 dist. &      0.0221 &     targ.1 goal dist. &         0.0142 &     herd.1  targ.4 dist. &      0.0173 &     targ.3 direction &      0.0166 &    herd.1  targ.1 dist. &      0.0175 \\
    5  &          targ.4 velocity & 0.0220 &     targ.1 direction &       0.0137 &    herd.1 accel. &       0.0160 &     herd.1  targ.4 dist. &      0.0158 &  targ.4 goal dist. &       0.0166 \\
    6  &     herd.1  targ.2 dist. &      0.0212 &     herd.1  targ.2 dist. &      0.0119 &     targ.2 direction &       0.0147 &    herd.2 targ.4 dist. &      0.0150 &   herd.velocity &       0.0148 \\
    7  &    herd.2 targ.4 dist. &      0.0200 &    herd.2 targ.3 dist. &      0.0117 &    herd.2 targ.4 dist. &  0.0142 &     herd.1  targ.2 dist. &  0.0138 &   herd.1  targ.3 dist. &      0.0144 \\
    8  &     herd.1  targ.3 dist. &      0.0199 &    herd.2 targ.4 dist. &      0.0115 &    targ.3 accel. &      0.0142 &     herd.1  targ.1 dist. &  0.0130 &    herd.1  targ.4 rel angle &      0.0142 \\
    9  &   h goal dist. &       0.0191 &     herd.1  targ.3 dist. &     0.0098 &     herd.1  targ.1 dist. &      0.0135 &   herd.2 targ.1 rel angle &      0.0116 &     herd.1  targ.2 dist. &      0.0133 \\
    10  &    targ.3 goal ang.&      0.0189 &        targ.3 goal ang. &     0.0096 &        targ.3 goal ang. &       0.0135 &    herd.2 targ.1 dist. &      0.0115 &          targ.4 velocity &      0.0132 \\
		\hline
		{} & \multicolumn{10}{c|}{$T_{hor} = 32$ (1280 ms) prediction horizon}\\
		\hline
		{} & \multicolumn{2}{l|}{Label 0} & \multicolumn{2}{l|}{Label 1} & \multicolumn{2}{l|}{Label 2} & \multicolumn{2}{l|}{Label 3} & \multicolumn{2}{l|}{Label 4}\\
		\hline   
		{} & Features & SHAP  &   Features & SHAP  &  Features & SHAP  &  Features & SHAP  &  Features & SHAP  \\
		{} &  &  values &    &  values &   &  values &   &  values &   &  values \\
		\hline
		{} & Features & SHAP  &   Features & SHAP  &  Features & SHAP  &  Features & SHAP  &  Features & SHAP  \\
		{} &  &  values &    &  values &   &  values &   &  values &   &  values \\
		\hline
		1  &     herd.1 targ.4 dist. &      0.0286 &    herd.2 targ.1 dist. &      0.0308 &  targ.2  goal dist. &       0.0364 &    herd.2 targ.3 dist. &      0.0212 &     herd.1 targ.4 dist. &      0.0261 \\
2  &     herd.1 targ.2 dist. &       0.0279 &     herd.1 targ.1 dist. &   0.0291 &     herd.1 targ.2 dist. &      0.0270 &     herd.1 targ.3 dist. &      0.0198 &    herd.2 targ.4 dist. &      0.0241 \\
3  &     herd.1 targ.1 dist. &       0.0264 &     herd.1 targ.4 dist. &   0.0283 &    herd.2 targ.2 dist. &      0.0245 &     targ.3  direction &     0.0182 &  targ.4  goal dist. &      0.0180 \\
4  &   h goal dist. &       0.0254 &  targ.2  goal dist. &   0.0273 &     herd.1 targ.4 dist. &      0.0233 &     herd.1 targ.2 dist. &      0.0177 &     herd.1 targ.1 dist. &      0.0143 \\
5  &          targ.4   velocity &       0.0250 &  targ.1 goal dist. &   0.0266 &   h goal dist. &      0.0215 &  targ.3  goal dist. &      0.0171 &     herd.1 targ.2 dist. &      0.0141 \\
6  &           h accel. &       0.0243 &     targ.1 direction &  0.0264 &     herd.1 targ.3 dist. &      0.0198 &     herd.1 targ.4 dist. &      0.0168 &    herd.1 targ.4 rel angle &      0.0137 \\
7  &    herd.1 targ.4 rel angle &       0.0240 &   h goal dist. &   0.0235 &     herd.1 targ.1 dist. &      0.0187 &     herd.1 targ.1 dist. &      0.0135 &    herd.2 targ.1 dist. &      0.0134 \\
8  &           h  velocity &       0.0227 &    herd.2 targ.3 dist. &   0.0197 &    herd.2 targ.3 dist. &      0.0169 &    herd.2 targ.1 dist. &      0.0134 &     herd.1 targ.3 dist. &      0.0126 \\
9  &        targ.3  goal ang. &      0.0226 &     herd.1 targ.3 dist. &   0.0184 &    herd.2 targ.1 dist. &      0.0163 &    herd.2 targ.4 dist. &      0.0128 &    herd.2 targ.2 dist. &      0.0120 \\
10  &     targ.1 direction &      0.0220 &     herd.1 targ.2 dist. &   0.0180 &  targ.1 goal dist. &      0.0160 &    herd.1 targ.4 rel angle &      0.0122 &    herd.2 targ.3 dist. &      0.0119 \\
		\hline
		
	\end{tabular}
	}
	\label{tab:SHAP_novice_1sec_balanced}
\end{table}

\begin{table}
	\centering
	\caption{Top 10 ranked features and corresponding SHAP values for each class predicted by the model trained on \textit{expert} pairs with a sequence $T_{seq} = 1 \mathrm{s}$ and model trained on balanced data.}
	\resizebox{\linewidth}{!}{%
	\begin{tabular}{|l|ll|ll|ll|ll|ll|}
			\hline
		{} & \multicolumn{10}{c|}{$T_{hor} = 16$ (640 ms) prediction horizon}\\
		\hline
		{} & \multicolumn{2}{l|}{Label 0} & \multicolumn{2}{l|}{Label 1} & \multicolumn{2}{l|}{Label 2} & \multicolumn{2}{l|}{Label 3} & \multicolumn{2}{l|}{Label 4}\\
		\hline     
		{} & Features & SHAP  &   Features & SHAP  &  Features & SHAP  &  Features & SHAP  &  Features & SHAP  \\
		{} &  &  values &    &  values &   &  values &   &  values &   &  values \\
		\hline
1  &    herd.2 targ.1 dist. &      0.0347 &    herd.2 targ.1 dist. &      0.0365 &    herd.2 targ.2 dist. &      0.0335 &    herd.2 targ.3 dist. &      0.0329 &    herd.2 targ.4 dist. &      0.0267 \\
2  &    herd.2 targ.2 dist. &      0.0313 &     herd.1 targ.1 dist. &      0.0209 &     herd.1 targ.1 dist. &      0.0203 &     herd.1 targ.1 dist. &      0.0183 &     targ.4  direction &     0.0145 \\
3  &    herd.2 targ.3 dist. &      0.0271 &    herd.2 targ.2 dist. &      0.0153 &  targ.2  goal dist. &      0.0190 &     herd.1 targ.3 dist. &      0.0172 &     herd.1 targ.4 dist. &      0.0145 \\
4  &     herd.1 targ.1 dist. &      0.0261 &     herd.1 targ.4 dist. &      0.0143 &     herd.1 targ.3 dist. &      0.0172 &     herd.1 targ.4 dist. &      0.0156 &    herd.2 targ.1 dist. &      0.0123 \\
5  &    herd.2 targ.4 dist. &      0.0259 &     targ.1 direction &     0.0140 &     herd.1 targ.2 dist. &      0.0155 &    herd.2 targ.2 dist. &      0.0139 &     herd.1 targ.3 dist. &      0.0123 \\
6  &     herd.1 targ.3 dist. &      0.0257 &     herd.1 targ.3 dist. &      0.0123 &    herd.2 targ.1 dist. &      0.0147 &    herd.2 targ.1 dist. &      0.0133 &     herd.1 targ.2 dist. &      0.0113 \\
7  &     herd.1 targ.4 dist. &      0.0241 &  targ.1 goal dist. &      0.0123 &     herd.1 targ.4 dist. &      0.0142 &     herd.1 targ.2 dist. &      0.0132 &    herd.2 targ.3 dist. &      0.0113 \\
8  &   h goal dist. &      0.0214 &     herd.1 targ.2 dist. &      0.0121 &   h goal dist. &      0.0129 &  targ.3  goal dist. &      0.0130 &    herd.2 targ.2 dist. &      0.0112 \\
9  &          targ.3   velocity &      0.0208 &    herd.2 targ.3 dist. &      0.0119 &    herd.2 targ.3 dist. &      0.0128 &     targ.2  direction &     0.0122 &  targ.4  goal dist. &      0.0111 \\
10  &     herd.1 targ.2 dist. &      0.0206 &    herd.2 targ.4 dist. &      0.0107 &    herd.2 targ.4 dist. &      0.0117 &     targ.3  direction &     0.0109 &     herd.1 targ.1 dist. &      0.0110 \\
		\hline
		{} & \multicolumn{10}{c|}{$T_{hor} = 32$ (1280 ms) prediction horizon}\\
		\hline
		{} & \multicolumn{2}{l|}{Label 0} & \multicolumn{2}{l|}{Label 1} & \multicolumn{2}{l|}{Label 2} & \multicolumn{2}{l|}{Label 3} & \multicolumn{2}{l|}{Label 4}\\
		\hline   
		{} & Features & SHAP  &   Features & SHAP  &  Features & SHAP  &  Features & SHAP  &  Features & SHAP  \\
		{} &  &  values &    &  values &   &  values &   &  values &   &  values \\
		\hline
		{} & Features & SHAP  &   Features & SHAP  &  Features & SHAP  &  Features & SHAP  &  Features & SHAP  \\
		{} &  &  values &    &  values &   &  values &   &  values &   &  values \\
		\hline
1  &    herd.2 targ.2 dist. &      0.0427 &    herd.2 targ.1 dist. &      0.0278 &    herd.2 targ.2 dist. &      0.0382 &    herd.2 targ.3 dist. &      0.0252 &    herd.2 targ.4 dist. &      0.0313 \\
2  &    herd.2 targ.4 dist. &      0.0349 &    herd.2 targ.2 dist. &      0.0173 &     herd.1 targ.4 dist. &      0.0187 &     herd.1 targ.4 dist. &      0.0211 &     herd.1 targ.4 dist. &      0.0178 \\
3  &     herd.1 targ.3 dist. &      0.0273 &     herd.1 targ.1 dist. &      0.0140 &     herd.1 targ.3 dist. &      0.0169 &    herd.2 targ.2 dist. &      0.0175 &  targ.4  goal dist. &      0.0167 \\
4  &     herd.1 targ.4 dist. &      0.0263 &  targ.1 goal dist. &      0.0140 &  targ.2  goal dist. &      0.0161 &     herd.1 targ.3 dist. &      0.0168 &    herd.2 targ.2 dist. &      0.0165 \\
5  &    herd.2 targ.1 dist. &      0.0243 &    herd.2 targ.4 dist. &      0.0139 &     herd.1 targ.1 dist. &      0.0151 &    herd.2 targ.4 dist. &      0.0165 &     herd.1 targ.2 dist. &      0.0159 \\
6  &  targ.4  goal dist. &      0.0226 &     herd.1 targ.3 dist. &      0.0129 &    herd.2 targ.4 dist. &      0.0142 &     herd.1 targ.1 dist. &      0.0158 &     herd.1 targ.3 dist. &      0.0158 \\
7  &     herd.1 targ.1 dist. &      0.0220 &     herd.1 targ.4 dist. &      0.0125 &     herd.1 targ.2 dist. &      0.0138 &    herd.2 targ.1 dist. &      0.0140 &     targ.4  direction &     0.0152 \\
8  &   h goal dist. &      0.0194 &     targ.1 direction &     0.0119 &   h goal dist. &      0.0134 &  targ.3  goal dist. &      0.0119 &     herd.1 targ.1 dist. &      0.0131 \\
9  &     herd.1 herd.2 dist. &      0.0185 &    herd.2 targ.3 dist. &      0.0105 &    herd.2 targ.3 dist. &      0.0120 &     herd.1 targ.2 dist. &      0.0115 &    herd.2 targ.1 dist. &      0.0121 \\
10  &    herd.2 targ.3 dist. &      0.0181 &     herd.1 targ.2 dist. &      0.0103 &    herd.2 targ.1 dist. &      0.0120 &   h goal dist. &      0.0103 &    herd.2 targ.3 dist. &      0.0099 \\
		\hline
		
	\end{tabular}
	}
	\label{tab:SHAP_expert_1sec_balanced}
\end{table}

\end{document}